\documentclass[journal,twoside,web]{ieeecolor}
\usepackage{jsen}
\usepackage{cite}
\usepackage{amsmath,amssymb,amsfonts}
\usepackage{algorithmic}
\usepackage{graphicx}
\usepackage{textcomp}
\usepackage{wrapfig}
\usepackage{flushend}
\usepackage{hyperref}
\usepackage{afterpage}
\usepackage{verbatim}

\def\BibTeX{{\rm B\kern-.05em{\sc i\kern-.025em b}\kern-.08em
    T\kern-.1667em\lower.7ex\hbox{E}\kern-.125emX}}
\definecolor{abstractbg}{rgb}{0.89804,0.94510,0.83137}
\begin{document}


\title{Soft Biomimetic Optical Tactile Sensing\\ with the TacTip: A Review}
\author{Nathan~F.~Lepora
\thanks{
This work was supported in part by Leverhulme Research Leadership Award on `A biomimetic forebrain for robot touch' (RL-2016-39). }
\thanks{N. Lepora is with the Department of Engineering Mathematics, Faculty of Engineering, University of Bristol and Bristol Robotics Laboratory, Bristol, UK. (e-mail: n.lepora@bristol.ac.uk). }}

\IEEEtitleabstractindextext{%
\fcolorbox{abstractbg}{abstractbg}{%
\begin{minipage}{\textwidth}%
\begin{wrapfigure}[15]{r}{3in}%
\includegraphics[width=2.75in]{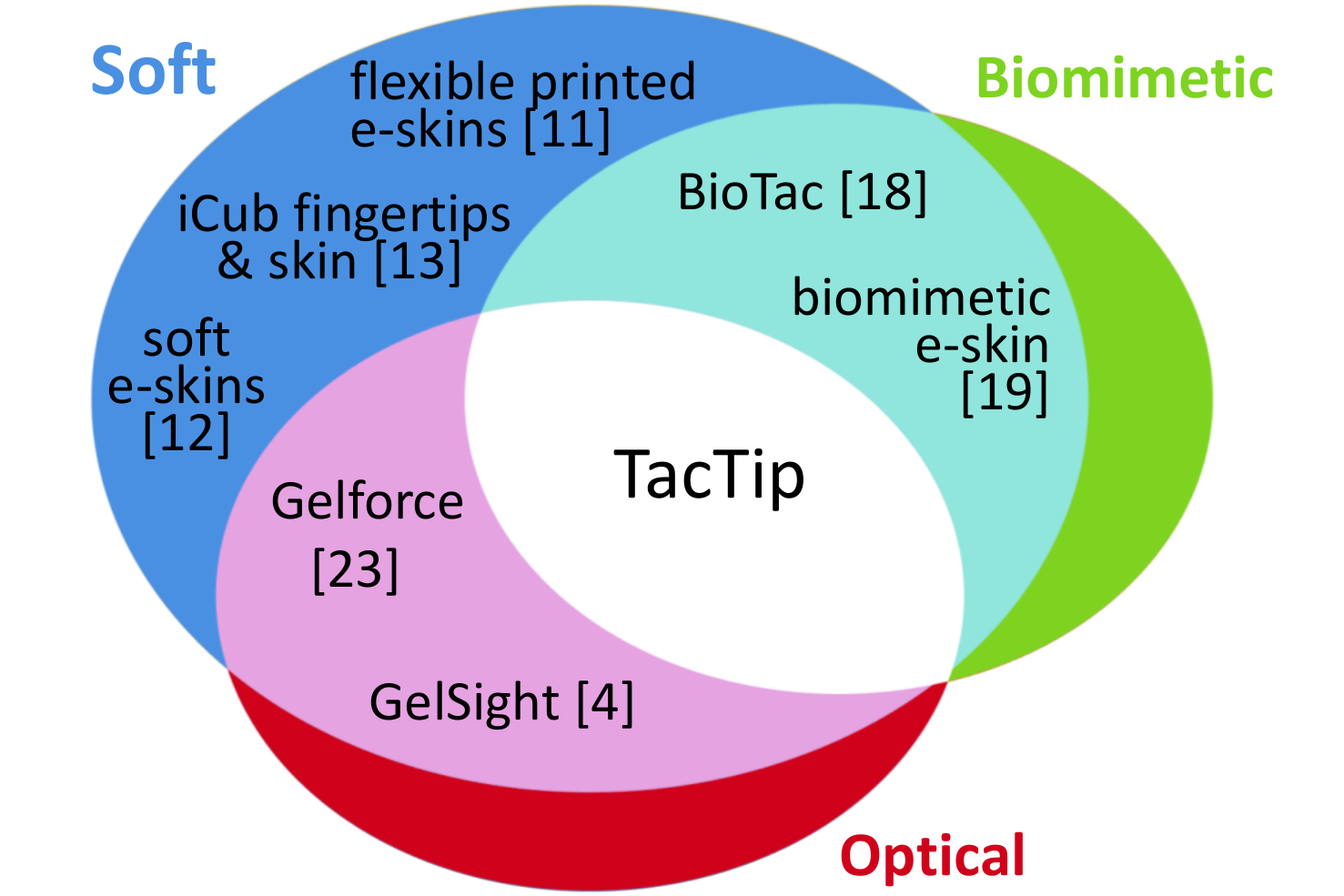}%
\end{wrapfigure}%
\begin{abstract}
Reproducing the capabilities of the human sense of touch in machines is an important step in enabling robot manipulation to have the ease of human dexterity. A combination of robotic technologies will be needed, including soft robotics, biomimetics and the high-resolution sensing offered by optical tactile sensors. This combination is considered here as a SoftBOT (Soft Biomimetic Optical Tactile) sensor. This article reviews the BRL TacTip as a prototypical example of such a sensor. Topics include the relation between artificial skin morphology and the transduction principles of human touch, the nature and benefits of tactile shear sensing, 3D printing for fabrication and integration into robot hands, the application of AI to tactile perception and control, and the recent step-change in capabilities due to deep learning. This review consolidates those advances from the past decade to indicate a path for robots to reach human-like dexterity.
\end{abstract}

\end{minipage}}}

\maketitle
 
\section{Introduction}
\label{sec:1}

As humans, we take our sense of touch for granted as we mostly use it in a subconscious way. Visual and auditory events grab our attention~\cite{oregan_why_2011} while our tactile sense continues unabated during everyday manual tasks and chores. However, touch is arguably the sense that most differentiates humans from other animals because the dexterous use of our hands relies on the intelligent use of tactile perception. With our hands we have invented technology, the hallmark of our species, spanning from the archaeological record of archaic~hominids to the modern era of advanced devices such as robots. 

Reproducing the capabilities of the human tactile sense in machines is an important step in enabling robotic hands to reach the dexterity of the human hand. As argued compellingly by historian Yuval Noah Harrari in his book ``Home Deus: A Brief History of Tomorrow", robot dexterity will have a profound impact on human society as machines become commonplace for physical labour~\cite{harari_homo_2016}. This revolution in robot dexterity is beginning in the logistics industry, but promises broader impacts in manufacturing, health, construction, food production, recycling, conservation and renewable energy.

A range of robotic technologies will be needed to reach human-like dexterity in machines. Soft robotics is needed for safe devices that adapt according to the intelligence embodied in their materials and design. Biomimetics provides the only known example of a general-purpose manipulation device: the human hand and our manual intelligence to use it effectively. Intelligent interaction requires a rich source of contact information, as offered by high-resolution optical tactile sensors such as the MIT GelSight~\cite{johnson_retrographic_2009,yuan_gelsight_2017} and BRL TacTip~\cite{chorley_development_2009,ward-cherrier_tactip_2018} with deep learning methods such as convolutional neural networks.

Here, this combination of technologies is termed SoftBOT sensing, for {\em Soft Biomimetic Optical Tactile Sensing}, encompassing SoftBOT sensors, hands and robotic systems. The BRL TacTip uniquely combines all of these features as a prototypical example of a SoftBOT sensor, and for this reason has led to a large body of research on robot dexterity. Therefore, this article reviews tactile sensing with the TacTip to consolidate the advances with this technology over the past decade and so indicate a path to human-like robot dexterity.



\section{Soft, Biomimetic and Optical Tactile Sensing}	
\label{sec:2}

To motivate this article, we define the terms `soft', `biomimetic' and `optical' in the context of tactile sensing, bearing in mind that a tactile sensor is a device that transduces deformation of a sensing surface into a signal containing information about the physical contact. As emphasised in past surveys~\cite{luo2017robotic,kappassov_tactile_2015,dahiya_tactile_2010,cutkosky_force_2008}, this information can be used to infer properties of the contacting object and its interaction with the tactile sensor.  

{\em Soft tactile sensors} rely on soft materials to elicit information from physical contact. While all tactile sensors use a deformable surface to sense, they range in material construction from flexible printed e-skins~\cite{khan_technologies_2015} to sensitive materials embedded within elastomeric substrates~\cite{wang_toward_2018} to compliant layers over electronic circuits that measure compression~\cite{schmitz_methods_2011}. Softness can have many benefits in the design and function of tactile sensors. For example, it is important that e-skins are flexible so they can conform their sensing surface to 3D objects and cover large areas of hard or soft actuated robots~\cite{dahiya_e-skin_2019}. 

{\em Soft biomimetic tactile sensors} are soft tactile sensors based on principles distilled from the study of biological systems~\cite{lepora_state_2013,pfeifer_challenges_2012}. There is a distinction between `blind copying', such as merely shaping a tactile sensor like a human fingertip, and true biomimicry, such as transferring the transduction principles of human skin into the design of an artificial sensor. Soft robots are often inspired by soft-bodied animals~\cite{kim_soft_2013}, so it is expected that biomimetic tactile sensors are usually soft. There are, however, many ways in which biological principles can motivate soft designs. One example is the inspiration for the Syntouch BioTac~\cite{wettels_biomimetic_2008} from the multi-modality of human touch to pressure, vibration and temperature. Another example is a biomimetic e-skin that measures local shear and normal forces by reproducing the hill-like structure of the dermal-epidermal boundary in human skin~\cite{boutry_hierarchically_2018}. 

{\em Soft optical tactile sensors} are soft tactile sensors that use light to view the deformation. Optical tactile sensors are often soft because they rely on viewing physical changes to the sensing surface, usually from inside the sensor with an internal light source. Optical touch sensing dates back to the mid-1960s~\cite{abad_visuotactile_2020}, with the earliest example relaying the view of a photoelastic skin via optic fibres to a TV camera whose signal was viewed remotely to teleoperate a robotic gripper~\cite{strickler_design_1966}. 

For soft camera-based tactile sensors, there are many proposals for imaging surface deformation, with categories including ~\cite{abad_visuotactile_2020,shimonomura_tactile_2019}: (i)~{\em soft marker-based optical tactile~sensors} that typically measure the (lateral) shear deformation of the sensing surface, such as the GelForce (2004) with its markers embedded in a supporting gel~\cite{kamiyama_evaluation_2004,kamiyama_vision-based_2005}; and (ii)~{\em soft reflection-based optical tactile sensors} that typically measure the (normal) indentation of the sensing surface, such as the GelSight (2009) which uses the surface shading from multiple internal light sources to infer a depth map via photometric stereo~\cite{johnson_retrographic_2009}. Combinations are also considered, such as by printing markers on the GelSight skin~\cite{yuan_measurement_2015} and by mixing coloured markers to indicate depth~\cite{lin_sensing_2019}. Recently, this area of research has been boosted by the remarkable advances in image processing using convolutional neural networks~\cite{krizhevsky_imagenet_2017}, which has led to many refinements of these designs~\cite{gomes_geltip_2020,lambeta_digit_2020,padmanabha_omnitact_2020,romero_soft_2020,lepora_towards_2021}. 

{\em Soft biomimetic optical tactile sensors} combine optical imaging with biological principles underlying the sense of touch in animals. A technology that offers both optical and biomimetic tactile sensing is the BRL TacTip (centre of figure in graphical abstract), proposed in 2009 as `a tactile sensor based on biologically-inspired edge encoding'~\cite{chorley_development_2009}. This article will review the biomimetic and optical principles of the TacTip as a prototypical example of a SoftBOT sensor. 


\section{Soft Biomimetic Optical Tactile Sensing:\\ The TacTip}
\label{sec:3}

\afterpage{%
\begin{figure*}[t!]
	\centering
	\begin{tabular}[b]{cc}
		{\bf (a) Skin physiology} & {\bf (b) Biomimetic tactile sensor} \\
		\includegraphics[width=0.3\textwidth,trim={0 125 0 120},clip]{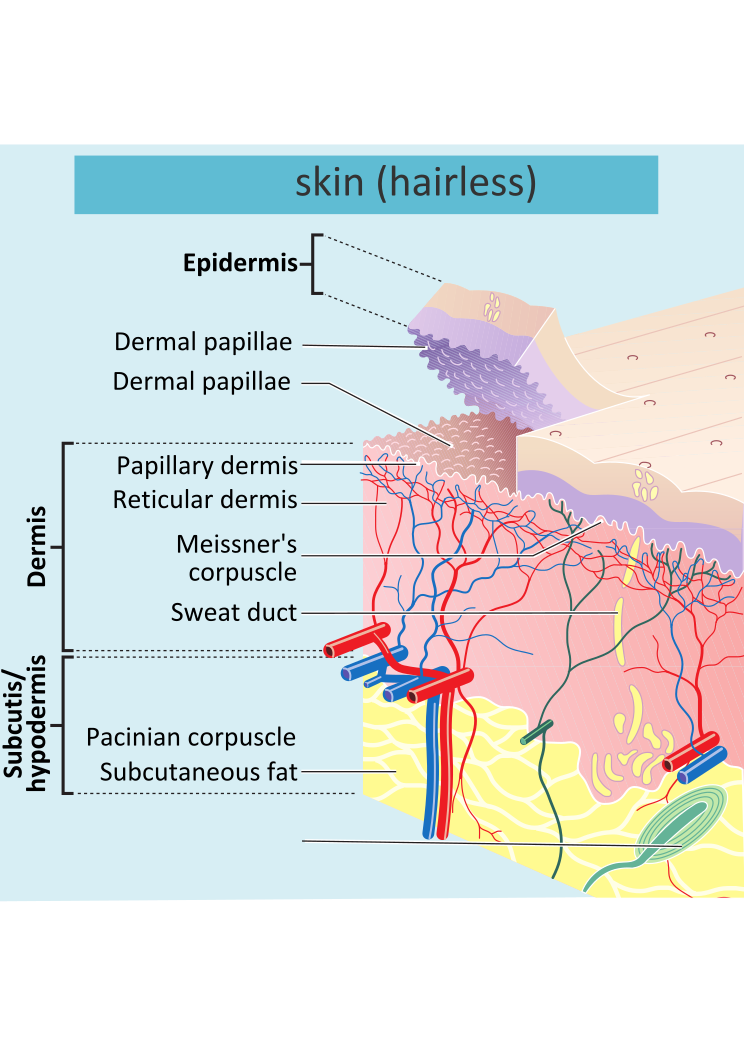} &
		\includegraphics[width=0.41\textwidth,trim={0 0 0 0},clip]{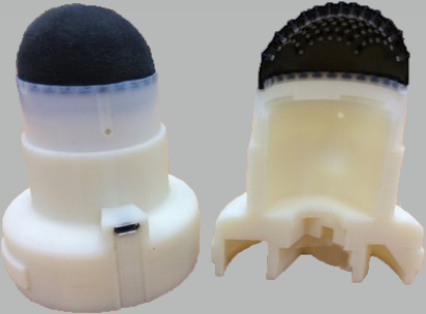}\\
		{\bf (c) Skin transduction} & {\bf (d) Biomimetic transduction} \\\textit{}
		\includegraphics[width=0.3\textwidth,trim={0 0 0 0},clip]{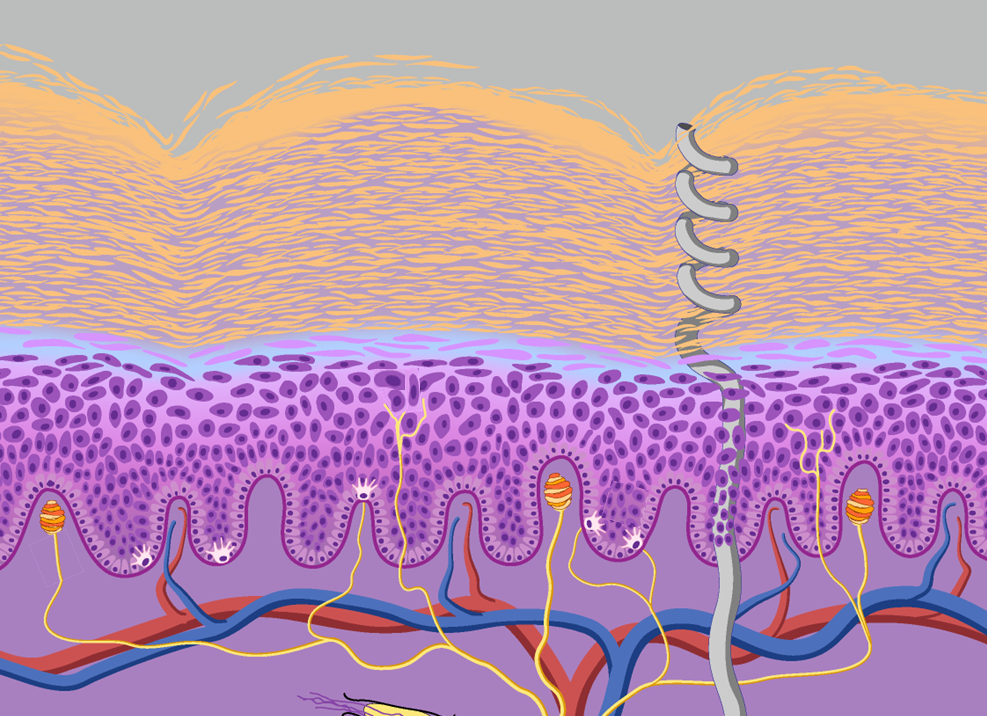} &
		\includegraphics[width=0.41\textwidth,trim={125 240 125 33},clip]{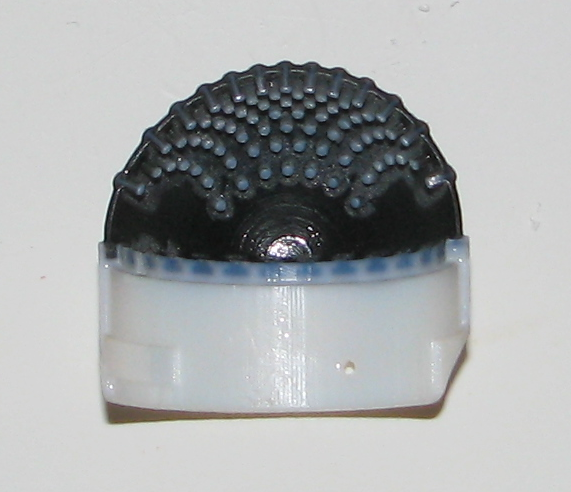}
	\end{tabular}
	\caption{Biomimetics of the TacTip. (a) Diagram of the layered morphology of hairless skin; (b) Cut-through of the 3D-printed BRL TacTip (2018); (c) Close-up of the interdigitation of dermis and epidermis, with sites of mechanoreceptors; (d) Close-up of a cut-through of the TacTip skin. The morphology of the artifical skin is based on natural skin. (Credits: Wikipedia, `Skin Layers', `Hegasy skin layers Receptors', CC By-SA License.)}
	\label{fig:1}
	\vspace{0em}
\end{figure*}

\begin{table*}[t]
	\centering 
	\begin{tabular}{@{}ccc@{}} 
		{\bf Neurophysiology} & {\bf Function} & {\bf Biomimetic counterpart} \\
		\hline
		epidermal ridges \& dermal papillae & transmits \& amplifies deformation of surface to mechanoreceptors & pins \& markers ~\cite{chorley_development_2009,ward-cherrier_tactip_2018} \\
		reticular dermis \& subcutaneous fat & soft structure \& compliance & elastomer gel~\cite{chorley_development_2009,ward-cherrier_tactip_2018}  \\
		SA-I mechanoreceptors (Merkel cells) & sense sustained skin deformation; perception of shape \& edges & pin displacements~\cite{lepora_biomimetic_2016,ward-cherrier_tactip_2018} \\ 
		RA-I mechanoreceptors (Meissner corpuscles) & sense transient skin movement; perception of flutter \& surface slip & pin velocities~\cite{james_slip_2018,james_slip_2020}  \\
		RA-II mechanoreceptors (Pacinian corpuscles) & vibration sensing; perception of surface texture & under investigation~\cite{winstone_tactip_2013,pestell_dual-modal_2018}  \\
		nociceptors (free nerve endings) & noxious touch & under investigation~\cite{pestell_dual-modal_2018} \\
		thermoceptors (free nerve endings) & temperature difference sensing & thermoactive skin~\cite{soter_multitip_2018}  \\
		overlapping sensitive receptive fields & hyperacuity & super-resolution~\cite{lepora_superresolution_2015} \\
	    epidermal ridges (fingerprint) & friction \& improved transduction; induces incipient slip & 3D-printed fingerprint~\cite{winstone_tactip_2013,cramphorn_addition_2017,james_slip_2020}  \\
		neural spiking & efficient signal encoding & event-based imaging~\cite{ward-cherrier_neurotac_2020,ward-cherrier_miniaturised_2020}  \\
	\end{tabular}
	\vspace{0em}
	\caption{Biomimetics of the TacTip, matching the neurophysiology and function.} 
	\label{table:1} 
	\vspace{-1em}
\end{table*}
}

The TacTip is a soft biomimetic optical tactile sensor that mimics the structure and function of the human fingertip (Figure~\ref{fig:1}). Human skin has an intricate morphology of layers, microstructures and sensory receptors that underlie its many functions, from protecting the body to sensing surface contact, temperature and injury~\cite{klatzky_touch_2003,abraira_sensory_2013,zimmerman_gentle_2014}. Like most other mammals, our skin is of two general types: hairy over much of our body and {\em glabrous} (hairless) on the sensitive underside of our fingers and toes, palms, soles of our feet, external genitalia, areolae, tongue, inner cheeks and lips.

The shallow layers of glabrous skin are structured to sense touch via the deformation of its surface. These upper layers comprise an outer {\em epidermis} over an inner {\em papillar dermis}, which interdigitate in a mesh of dermal {\em papillae} and epidermal {\em intermediate ridges} (Figures~\ref{fig:1}a,c). This 3D structure transmits shear and normal deformation of the skin surface into the displacement of sensory mechanoreceptors within the dermal-epidermal interface. Thus, the dermal papillae and intermediate ridges are considered to act as a mechanical amplifier of skin deformation into mechanoreceptor activity~\cite{cauna_nature_1954}. 

The TacTip design is based on the shallow layers of glabrous skin~\cite{chorley_development_2009,ward-cherrier_tactip_2018}. It has an outer biomimetic epidermis made from a rubber-like material over a soft inner biomimetic dermis made from an elastomer gel (Figures~\ref{fig:1}b,d). These two materials interdigitate in a mesh of biomimetic intermediate ridges and dermal papillae, comprising stiff inner nodular pins that extend under the biomimetic epidermis into the soft gel. This structure amplifies surface deformation of the skin into lateral movement of visible markers on the pin tips.

In human skin, two types of sensory mechanoreceptor are embedded within the dermal-epidermal interface: {\em Merkel cells} on the intermediate ridges and {\em Meissner corpuscles} within the dermal papillae (Figure~\ref{fig:1}c). Merkel cells respond to skin deformation, activating {\em slowly-adapting} (SA) neurons that fire during sustained contact, such as when perceiving shapes or edges. Meissner corpuscles respond to skin movement, activating {\em rapidly-adapting} (RA) neurons that fire during changes of contact, such as when perceiving surface slip and flutter. These sensory receptors work together to convey information about the tangible aspects of touch~\cite{saal_touch_2014}.

The biomimetic counterparts to these sensory mechanoreceptors are the markers on the pin tips, which can be imaged through a transparent gel forming the dermis~(Table~\ref{table:1}). The marker displacements from rest are analogous to the SA activity from the Merkel cells because they indicate sustained deformation of the sensing surface~\cite{lepora_superresolution_2015} to enable accurate shape and edge recognition~\cite{lepora_biomimetic_2016,lepora_exploratory_2017}. Likewise, the counterpart of RA mechanoreceptor activity from the Meissner corpuscles are the marker velocities, because these comprise a transient signal that indicates skin motion. Accordingly, the marker velocities enable accurate slip detection~\cite{james_slip_2018,james_slip_2020}. Together, these two signals -- the marker displacements and velocities -- give information about the sustained deformation and transient motion of the sensing surface. 

The deeper layers of glabrous skin also have biomimetic counterparts in the TacTip design~(Table~\ref{table:1}). Skin maintains its compliance from the deep {\em reticular dermis} and {\em subcutaneous fat} (Figure~\ref{fig:1}a), whose biomimetic counterpart is the deeper elastomer gel held in place by a transparent acrylic cap. Within the deep skin layers, a population of RA mechanoreceptors called {\em Pacinian corpuscles} responds to high-frequency vibration (peak sensitivity $\sim$250\,Hz). A partial mimicry of their function can be attained by using the TacTip with a high frame-rate (kHz) camera~\cite{winstone_tactip_2013,pestell_dual-modal_2018}; however, questions remain about whether this approach to vibration sensing is effective or even biomimetic, since it images fast pin movement rather than vibration in the deeper gel. In our view, a biomimetic counterpart of the vibration sense would be to embed a pressure sensor in the gel of the TacTip, like the vibration modality of the BioTac~\cite{wettels_biomimetic_2008}. Other tactile sensing modalities can also be included, such as temperature sensing by using a thermoactive smart material for the outer TacTip skin, which is imaged as a background to the markers~\cite{soter_multitip_2018}.

A consequence of the biomimetic design of the TacTip is that other properties of human perception emerge. An important aspect of human tactile perception is {\em hyperacuity}: a capacity to discriminate extended spatial features to a sub-millimetre acuity that is finer than the millimetre-scale spacing between mechanoreceptors~\cite{loomis_investigation_1980}. The TacTip also exhibits tactile hyperacuity, with a sub-millimetre capacity for spatial discrimination that is finer than its millimetre-scale pin spacing~\cite{lepora_superresolution_2015}. Fundamentally, the hyperacuity arises because both the biological and artificial tactile senses are comprised of arrays of overlapping, broad but sensitive receptive fields. This structure enables spatial interpolation over neighbouring receptors, which is analogous to a well-known technique in optical imaging known as super-resolution~\cite{lepora_tactile_2015}.

Perhaps surprisingly, the role of the human fingerprint in the sense of touch is still being investigated~\cite{yum_fingerprint_2020} after two centuries of study. A 3D-printed fingerprint can be reproduced in the TacTip as raised bumps~\cite{winstone_tactip_2013,cramphorn_addition_2017} or concentric raised rings over the papillae~\cite{james_tactile_2020}. Benefits of a biomimetic fingerprint include increased sensitivity to texture~\cite{winstone_tactip_2013} and spatial localisation~\cite{cramphorn_addition_2017}. A ringed biomimetic fingerprint can also induce incipient slip~\cite{james_tactile_2020}, where a local region of skin slips in detectable way before the grip slips~\cite{james_slip_2018,james_slip_2020}. 

Finally, another direction to take biomimetic optical tactile sensing is to adopt neuromorphic transduction using trains of `spike' events~\cite{chortos_pursuing_2016,yi_biomimetic_2018}. Biological nervous systems use event-based signalling to represent sensory information in an efficient and temporally-precise way, which neuromorphic engineering aims to reproduce artificially~\cite{mead_neuromorphic_1990}. A benefit of optical tactile sensing with the TacTip is that  an event-based camera can be used to implement a transduction mechanism that has both a biomimetic skin design and is neuromorphic from the camera~\cite{ward-cherrier_neurotac_2020,ward-cherrier_miniaturised_2020}. Overall, neuromorphic computation offers a new paradigm for robot touch that complements the capabilities of soft biomimetic optical tactile sensors.


\begin{figure*}[h!]
	\centering
	\renewcommand{\tabcolsep}{2pt}
	\begin{tabular}[b]{ccccc}
	\multicolumn{5}{c}{\bf (a) Single tactile image processed into shear-strain field (coloured by strain)} \\
	{Tactile image \& setup} & {{$x$-shears}} & {$y$-shears} & {Shear magnitude} & {Voronoi areas} \\
	\includegraphics[width=0.19\textwidth,trim={0 0 0 0},clip]{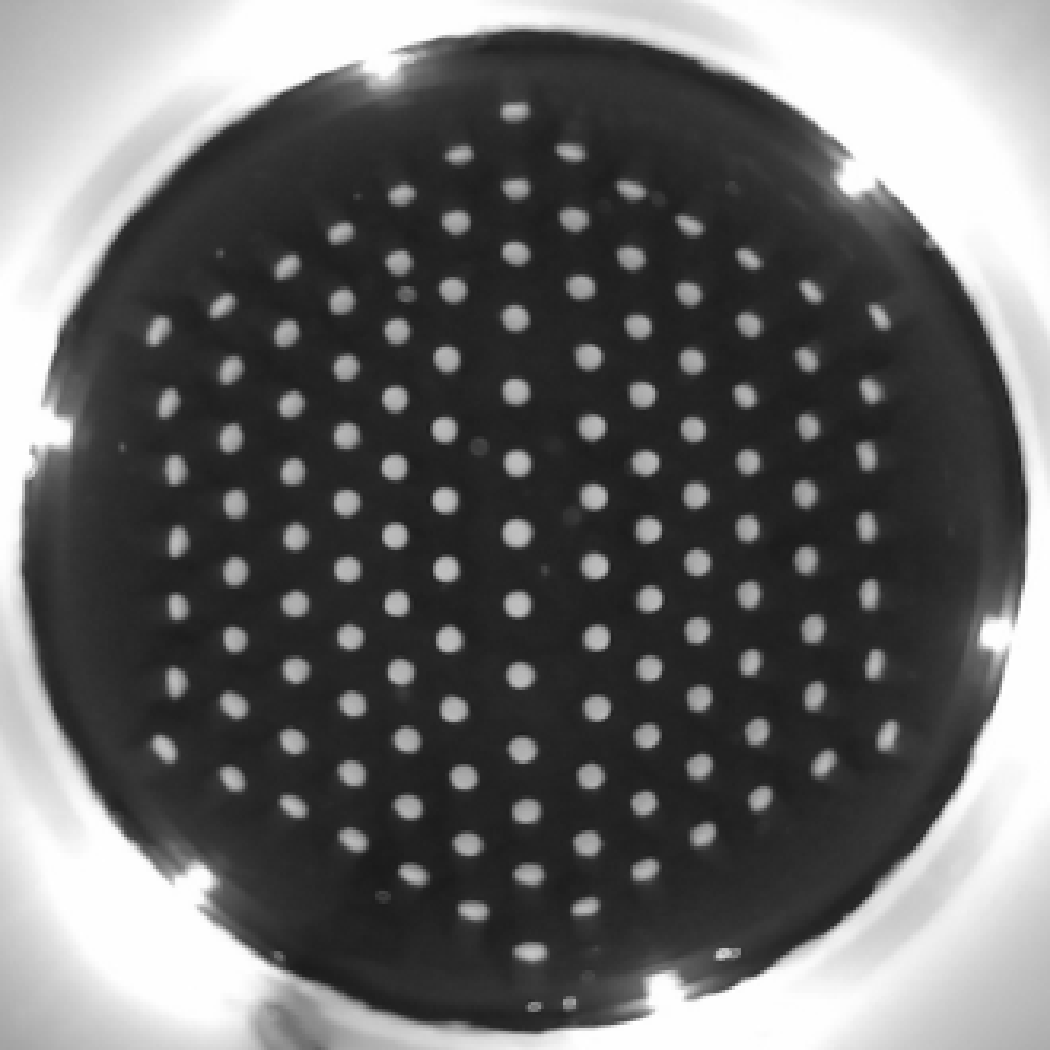}&
	\includegraphics[width=0.19\textwidth,trim={30 30 20 20},clip]{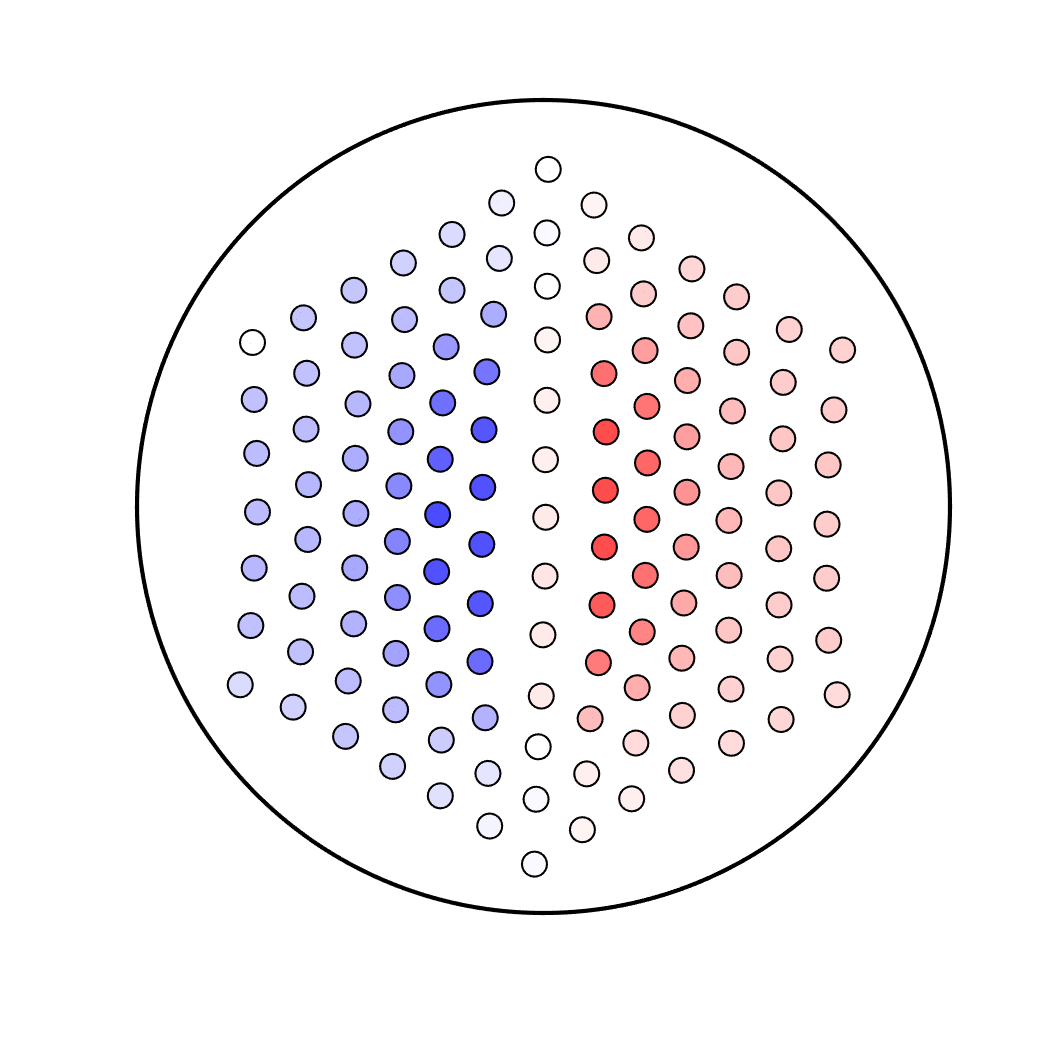}&		
	\includegraphics[width=0.19\textwidth,trim={30 30 20 20},clip]{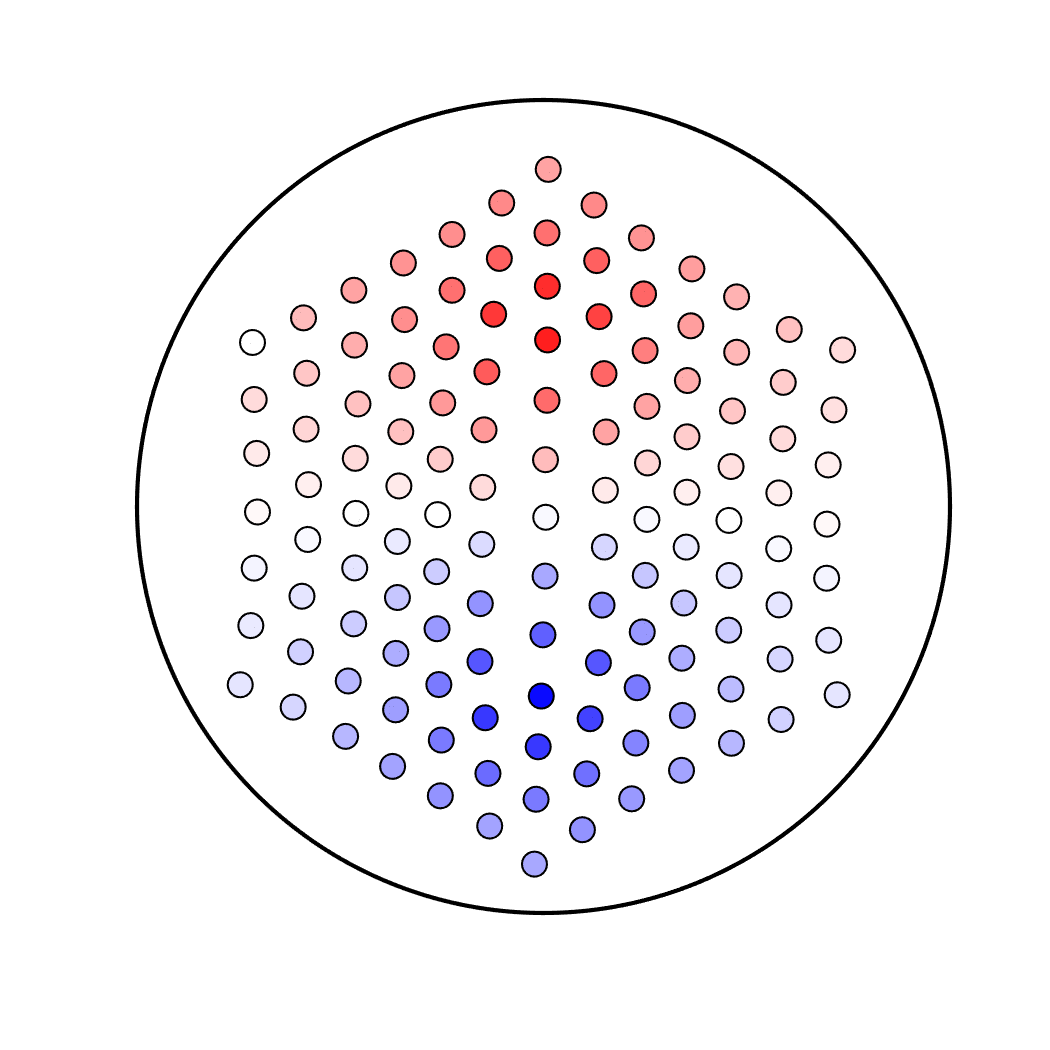}&
	\includegraphics[width=0.19\textwidth,trim={30 30 20 20},clip]{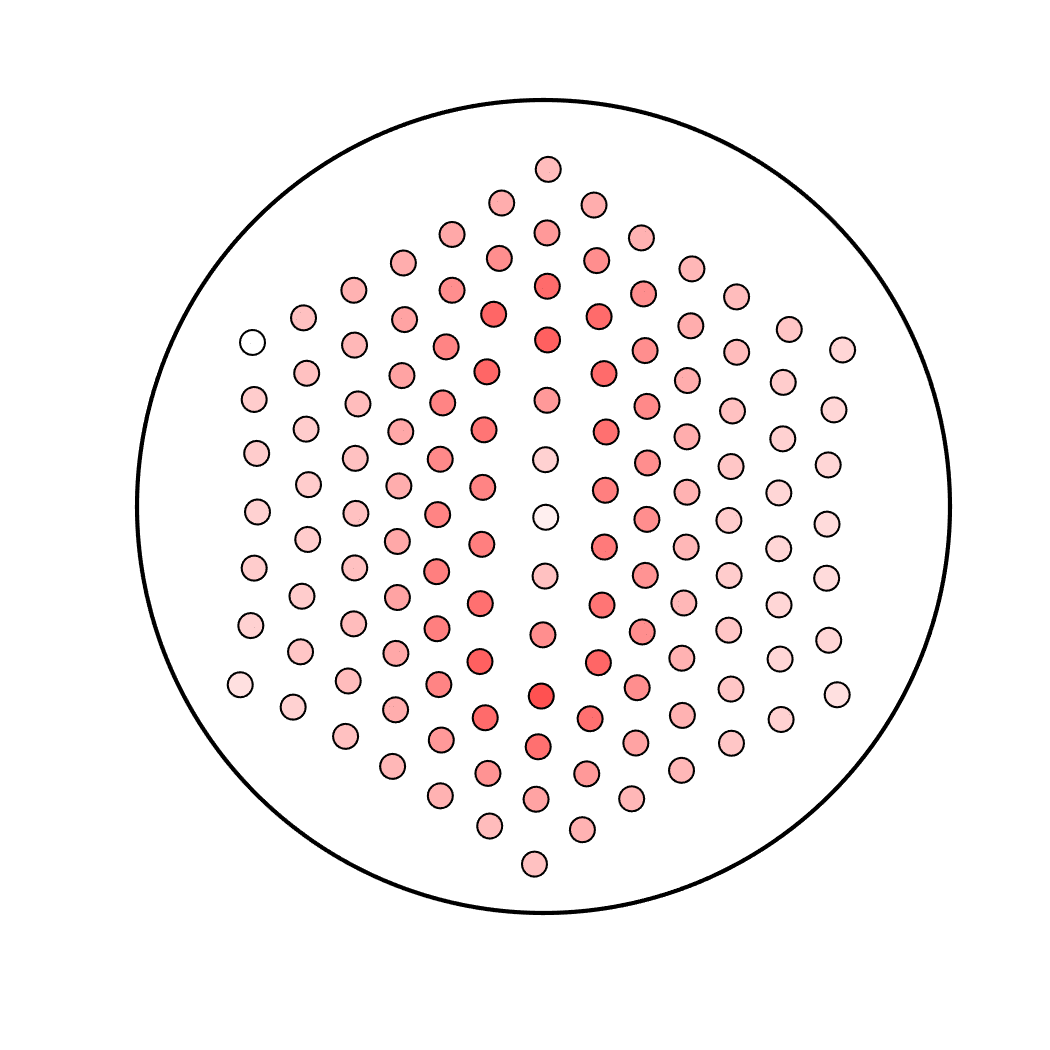}& 
	\includegraphics[width=0.19\textwidth,trim={30 30 20 20},clip]{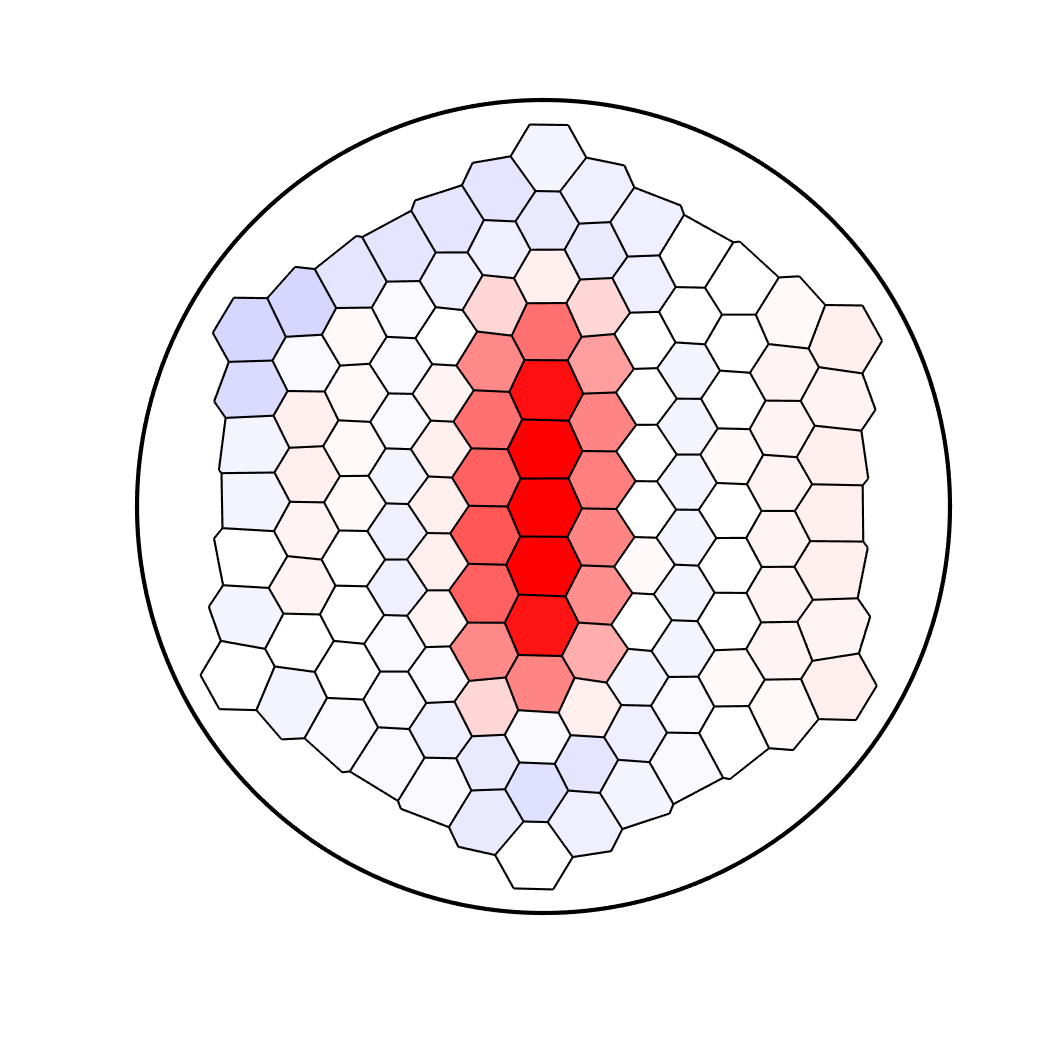} \\
	\includegraphics[width=0.19\textwidth,trim={30 30 30 30},clip]{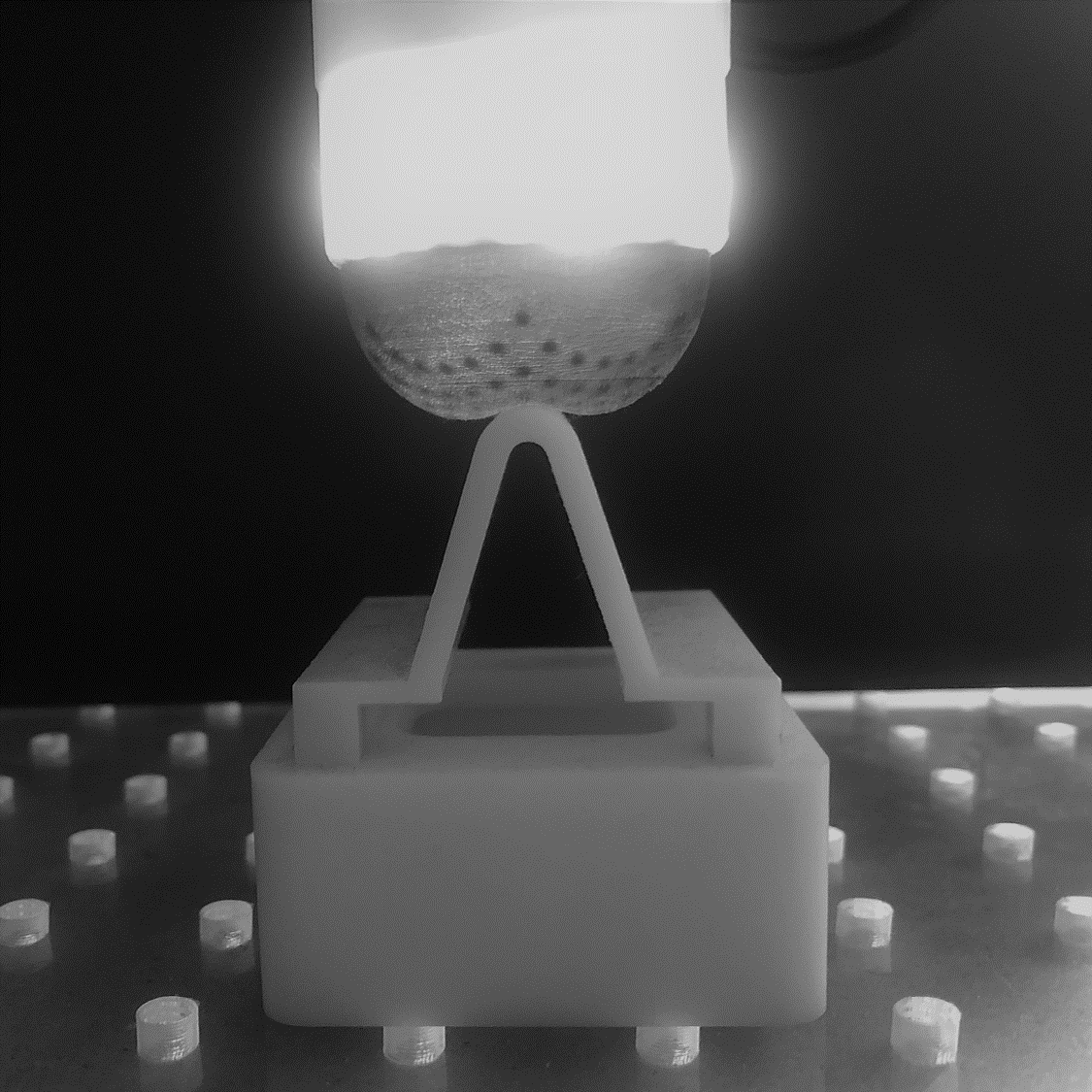}&
	\includegraphics[width=0.19\textwidth,trim={30 30 20 20},clip]{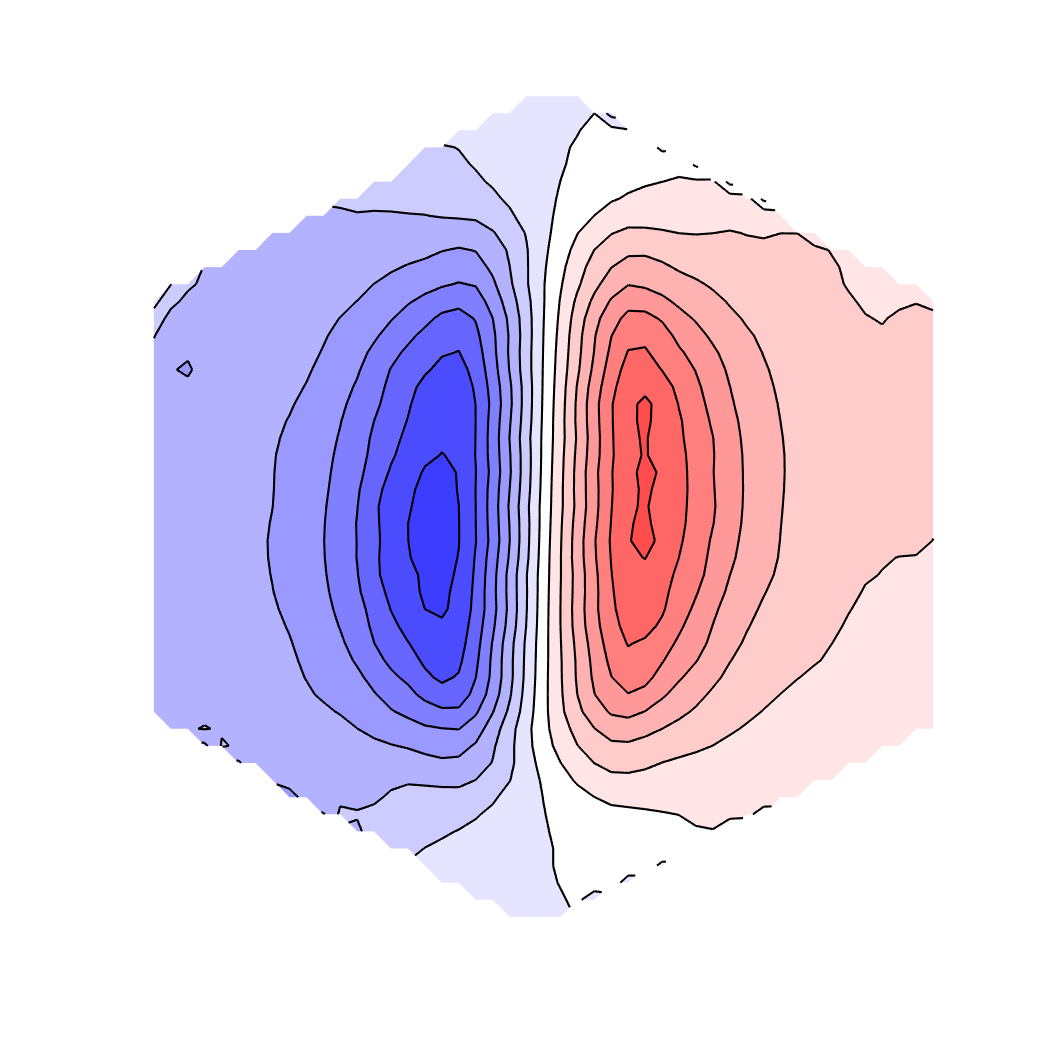}&		
	\includegraphics[width=0.19\textwidth,trim={30 30 20 20},clip]{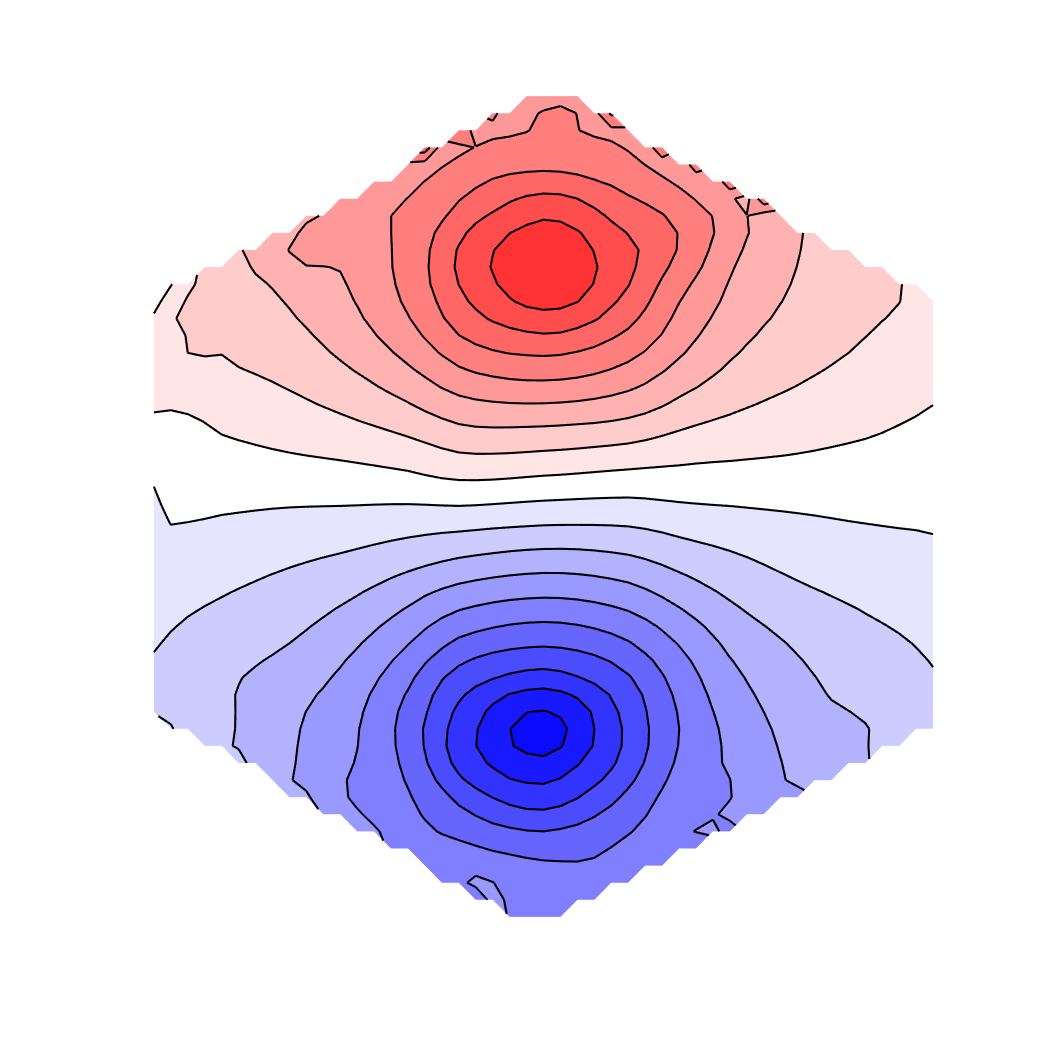}&
	\includegraphics[width=0.19\textwidth,trim={30 30 20 20},clip]{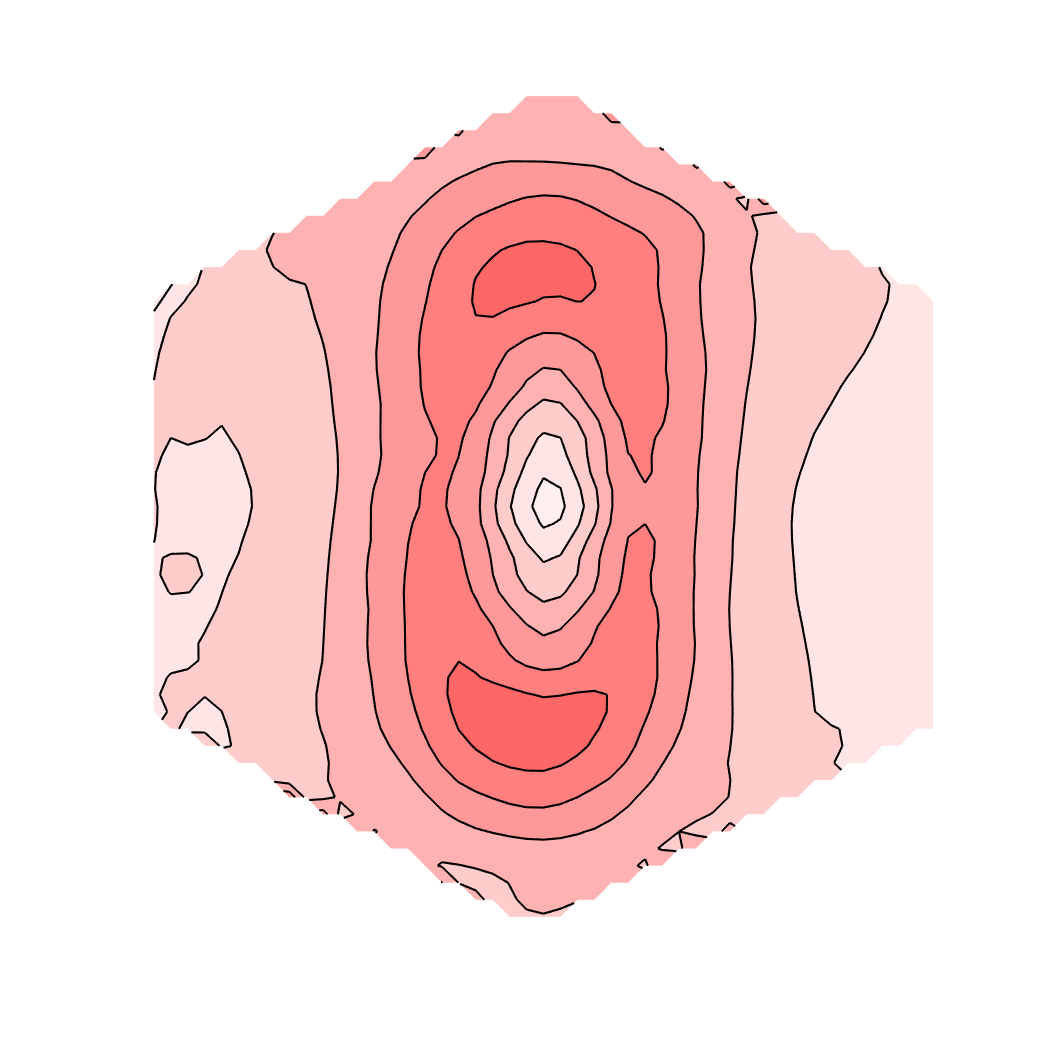}&
	\includegraphics[width=0.19\textwidth,trim={30 30 20 20},clip]{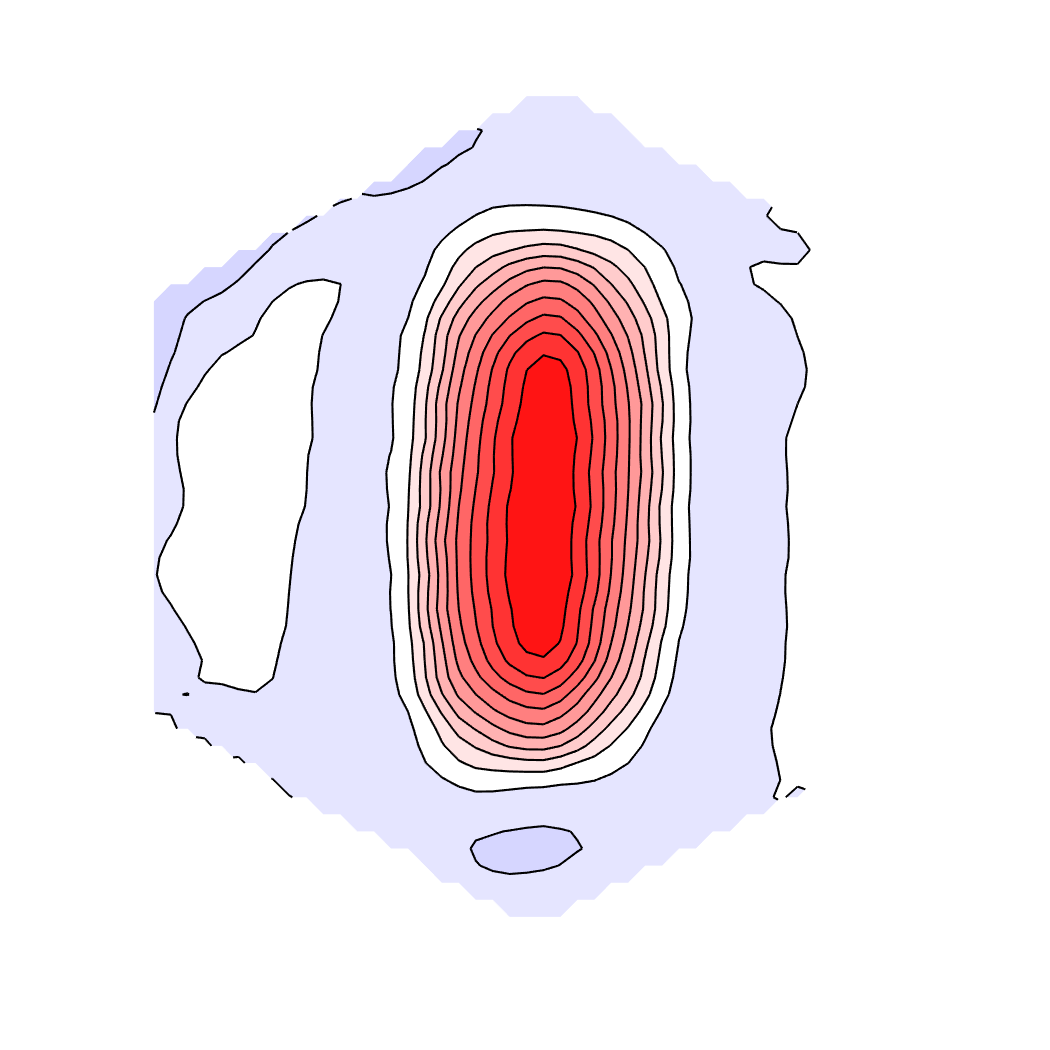} \\
	&&&&\\
	\end{tabular}\\
	\vspace{.5em}
	\renewcommand{\tabcolsep}{3pt}
	\begin{tabular}[b]{ccc}
		\multicolumn{3}{c}{\bf (b) Biomimetic signal transduction} \\
		{Skin indentation} & {Levering of pins} & {\ \ \ \ Indentation and shear strain}  \\
		\includegraphics[width=0.32\textwidth,trim={100 10 110 0},clip]{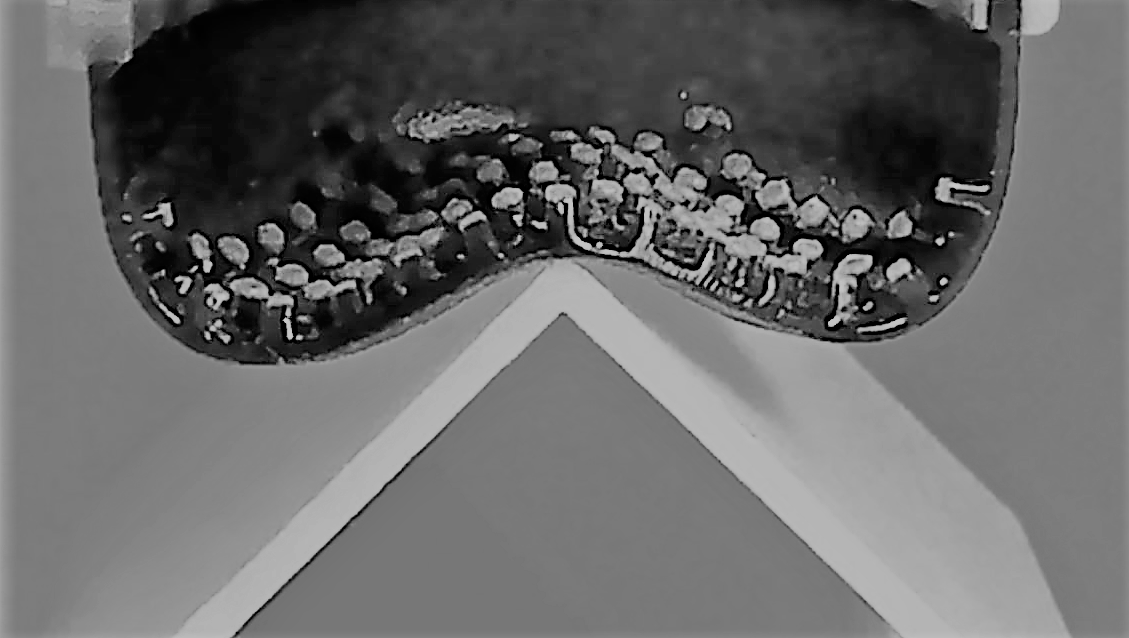}&
		\includegraphics[width=0.32\textwidth,trim={100 625 200 3},clip]{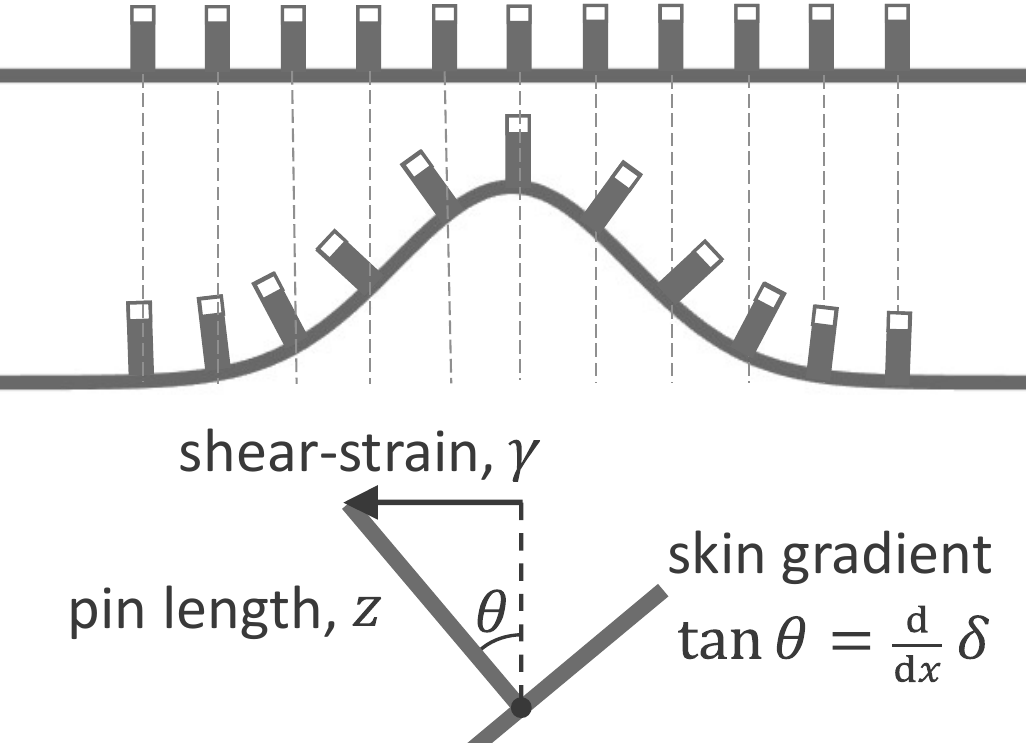}&
		\includegraphics[width=0.32\textwidth,trim={0 0 0 0},clip]{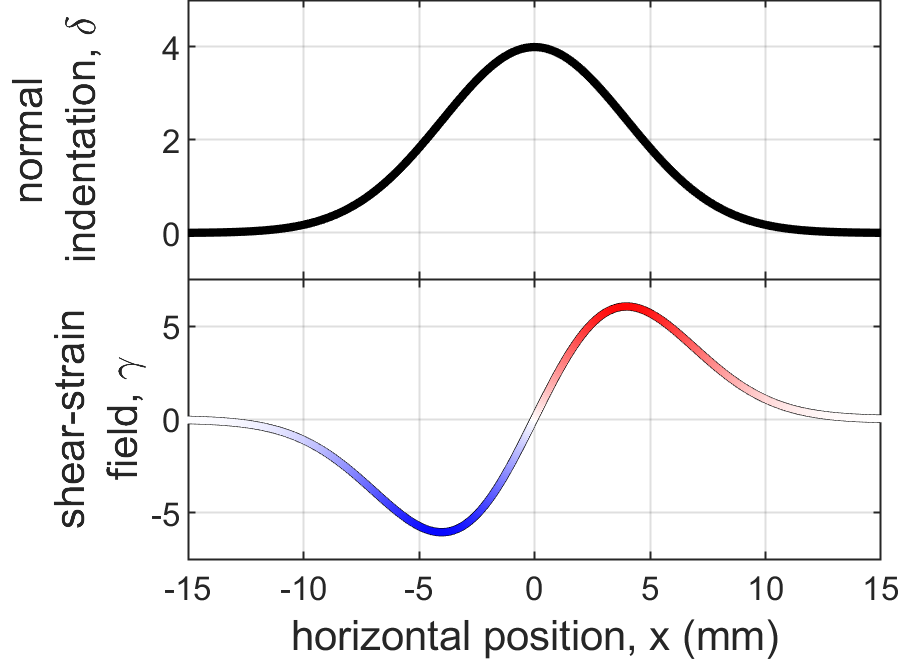}\\
		&&\\
	\end{tabular}\\
	\vspace{.5em}
	\renewcommand{\tabcolsep}{2pt}
	\begin{tabular}[b]{ccccc}
	\multicolumn{5}{c}{\bf (c) Time series of tactile images processed into shear strains (coloured by marker)} \\
	{Sensor position} & {$x$-shear strains} & {$y$-shear strains} & {Shear-strains} & {Voronoi areas} \\
	\includegraphics[width=0.19\textwidth,trim={0 0 15 15},clip]{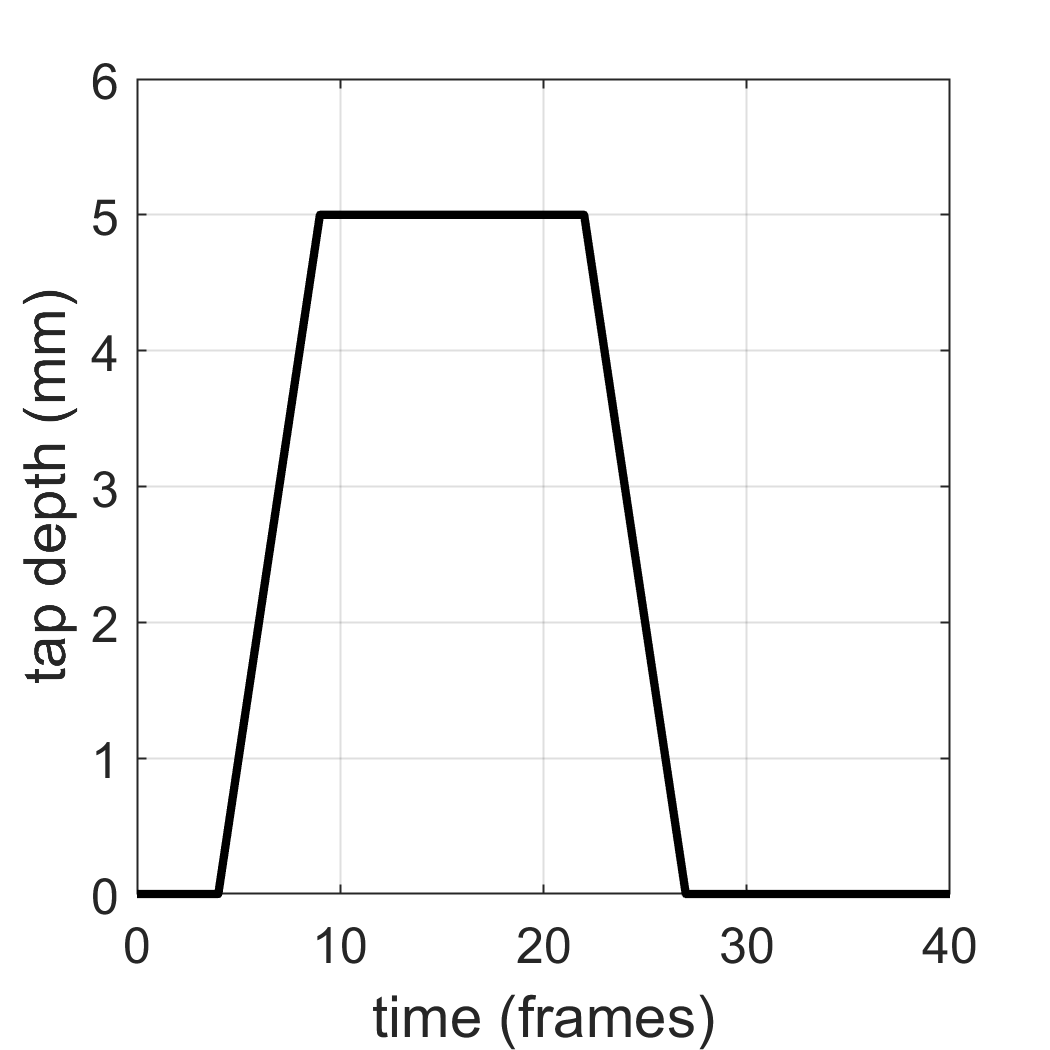}&
	\includegraphics[width=0.19\textwidth,trim={0 0 15 15},clip]{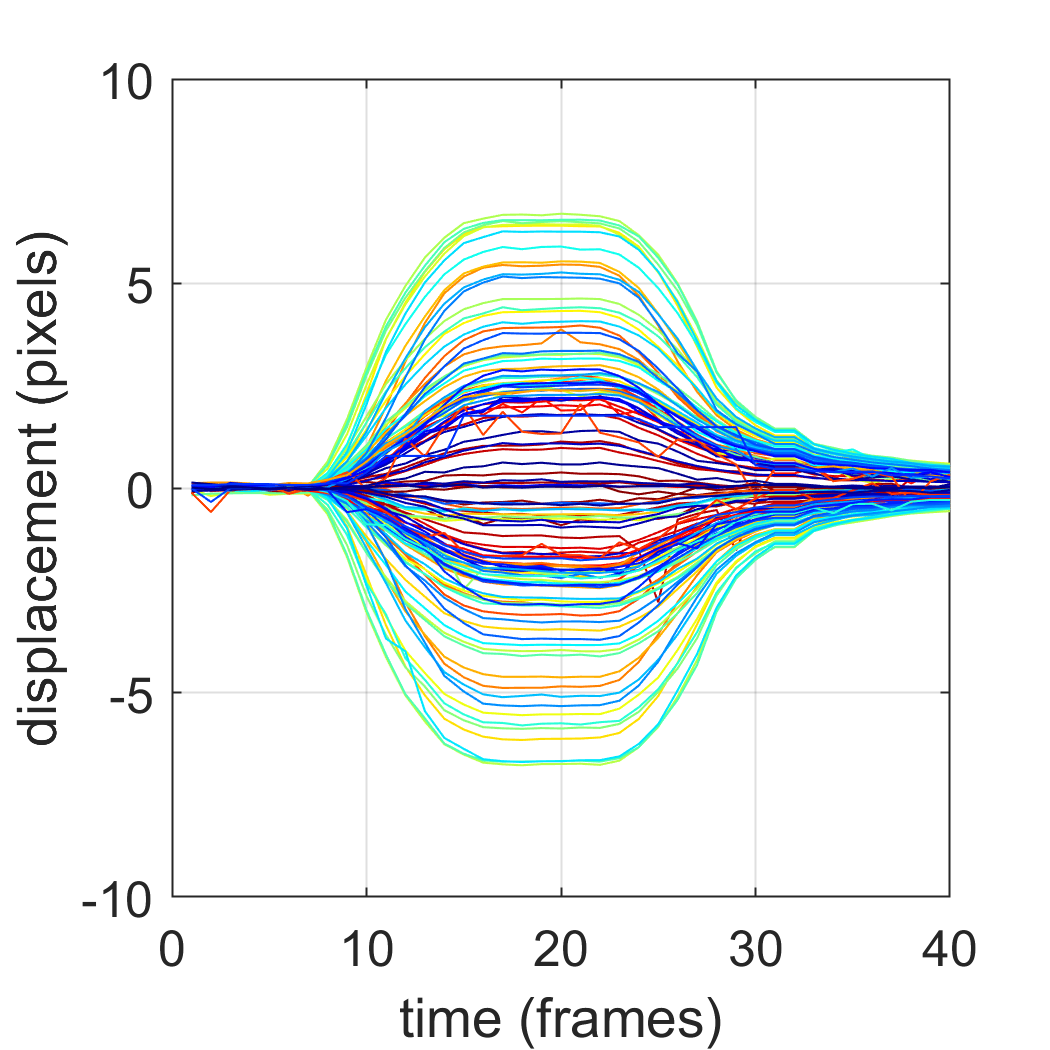}&
	\includegraphics[width=0.19\textwidth,trim={0 0 15 15},clip]{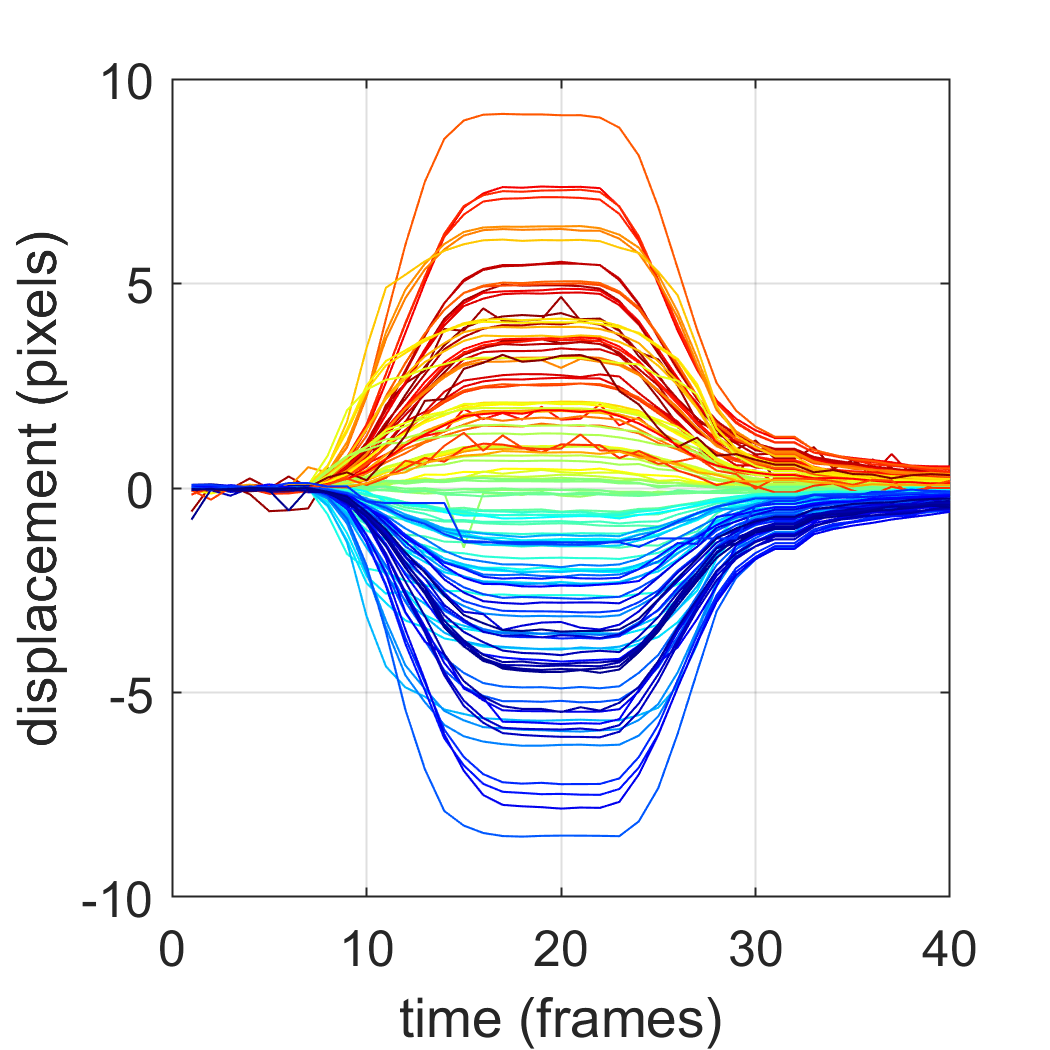}&
	\includegraphics[width=0.19\textwidth,trim={0 0 15 15},clip]{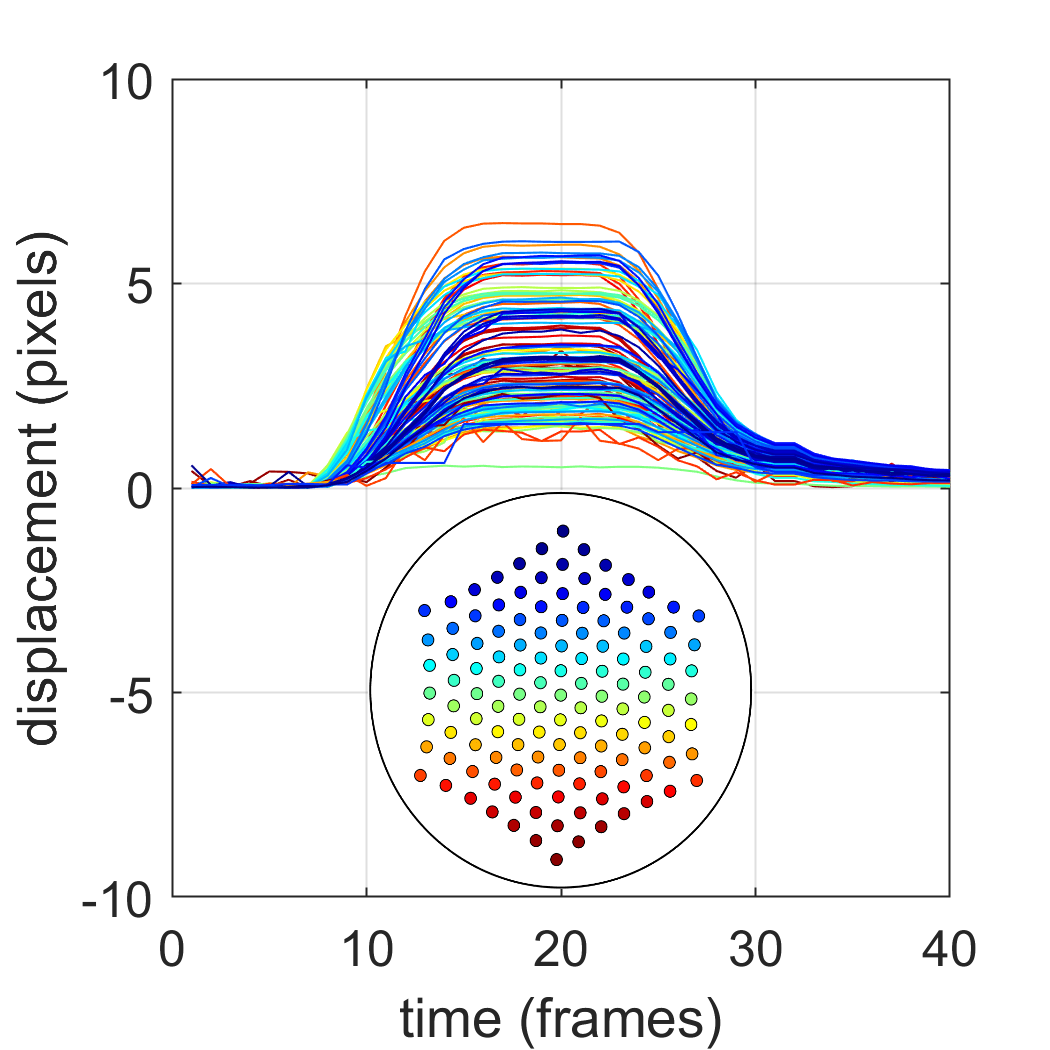}&
	\includegraphics[width=0.19\textwidth,trim={0 0 15 15},clip]{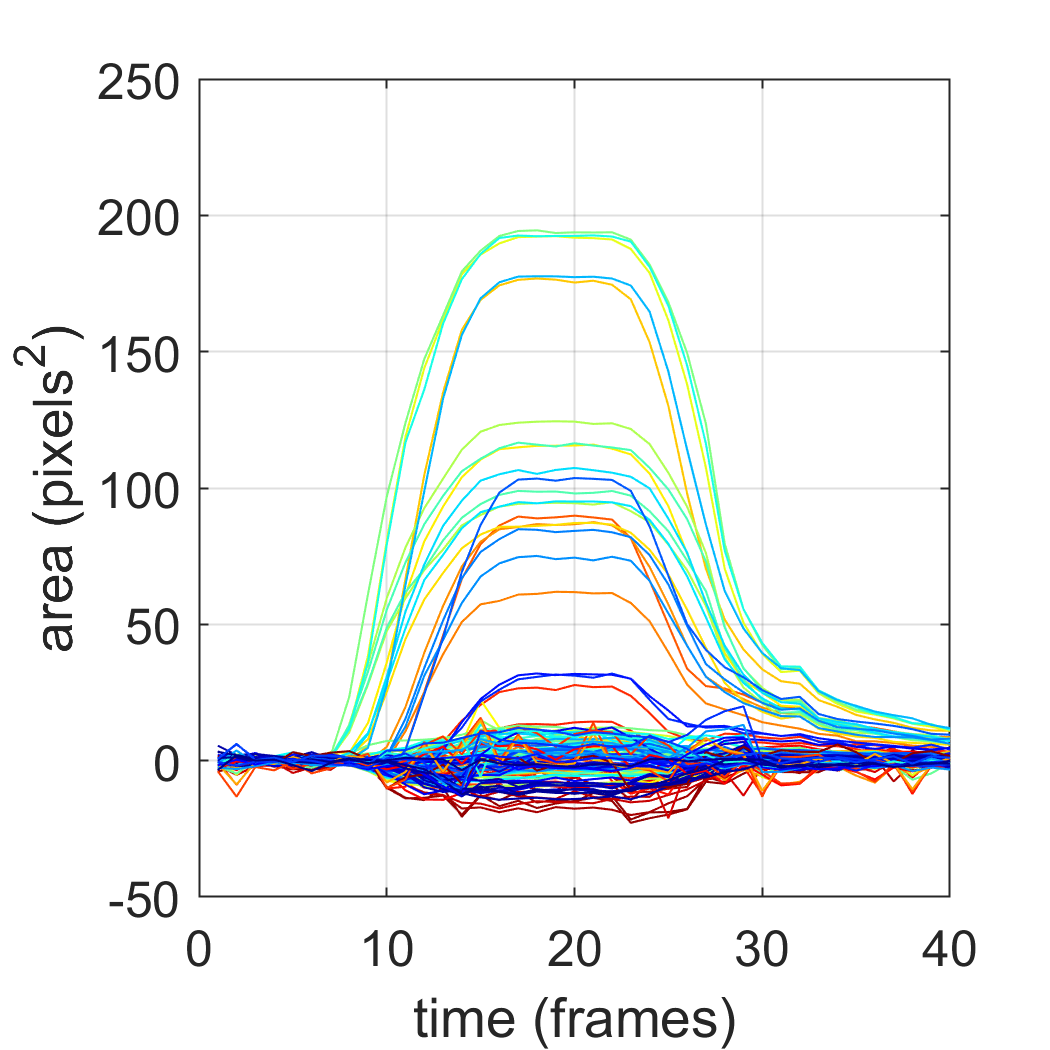} \\
	{Sensor velocity} & {$x$-shear-velocities} & {$y$-shear-velocities} & {Shear-velocities} & {Voronoi area changes} \\
	\includegraphics[width=0.19\textwidth,trim={0 0 15 15},clip]{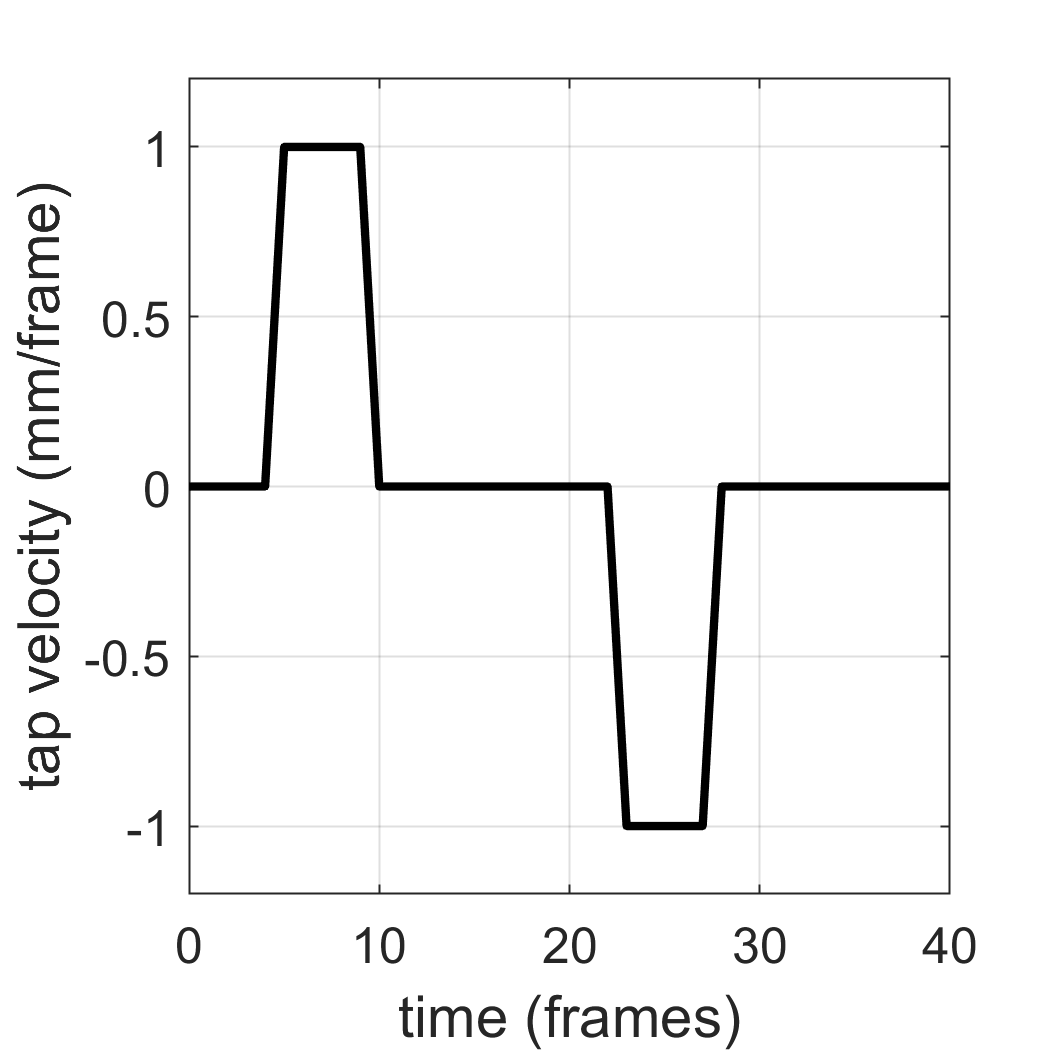}&
	\includegraphics[width=0.19\textwidth,trim={0 0 15 15},clip]{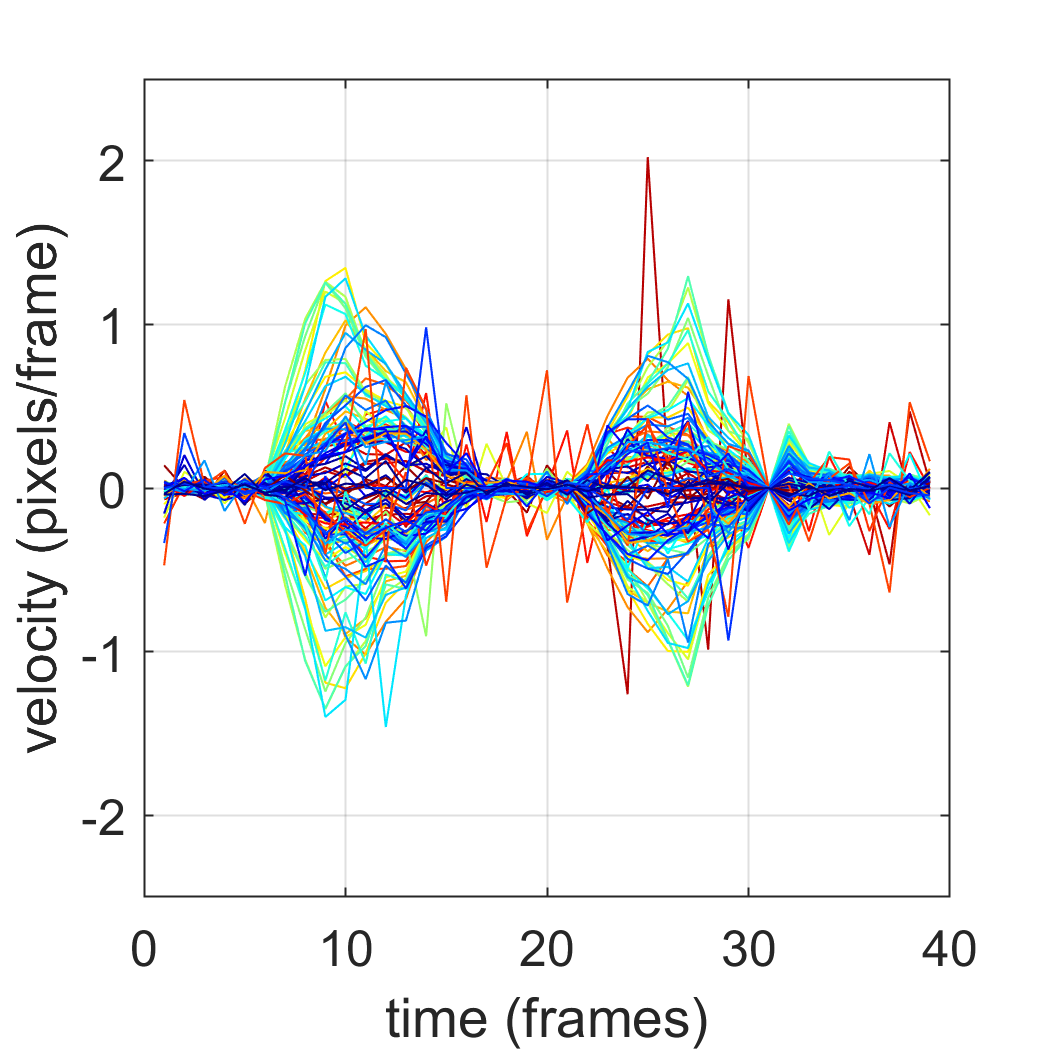}&
	\includegraphics[width=0.19\textwidth,trim={0 0 15 15},clip]{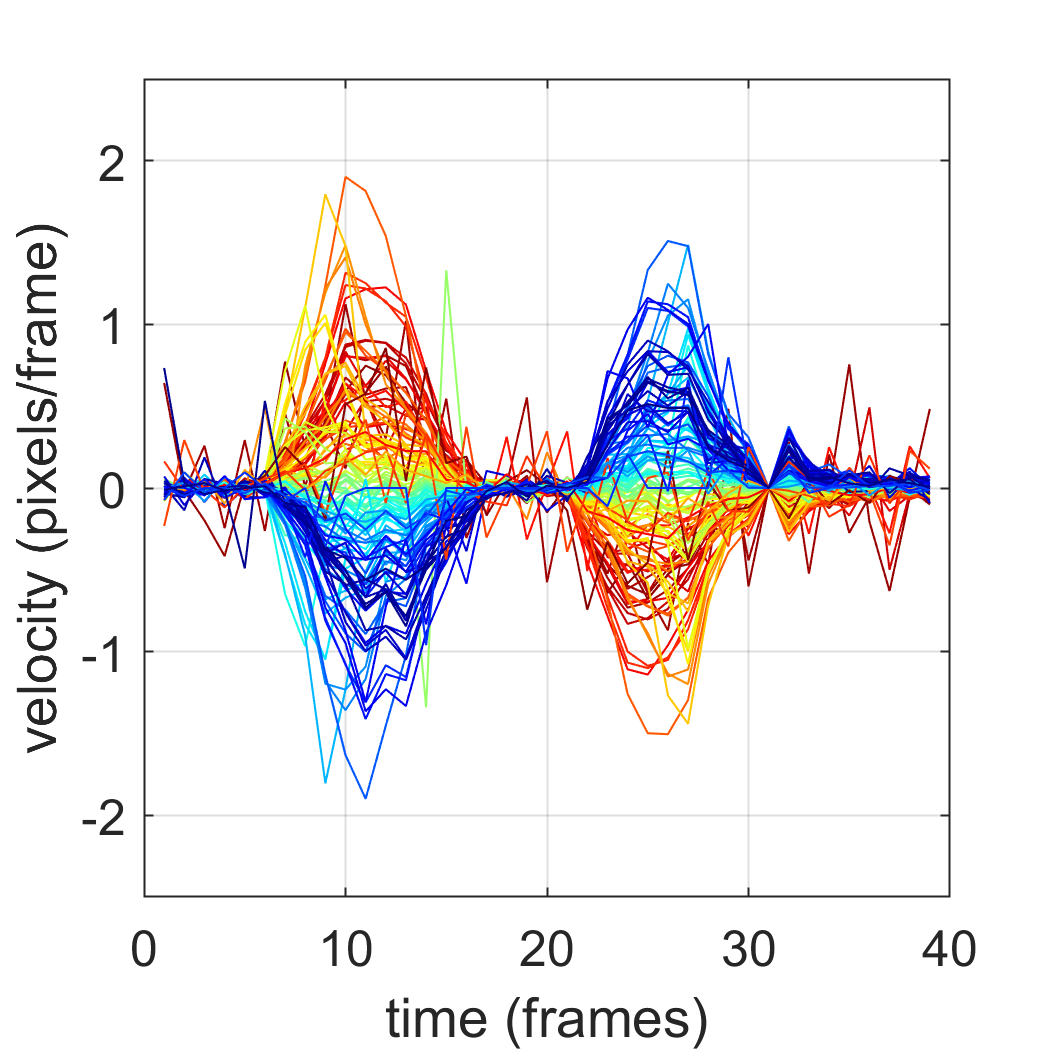}&
	\includegraphics[width=0.19\textwidth,trim={0 0 15 15},clip]{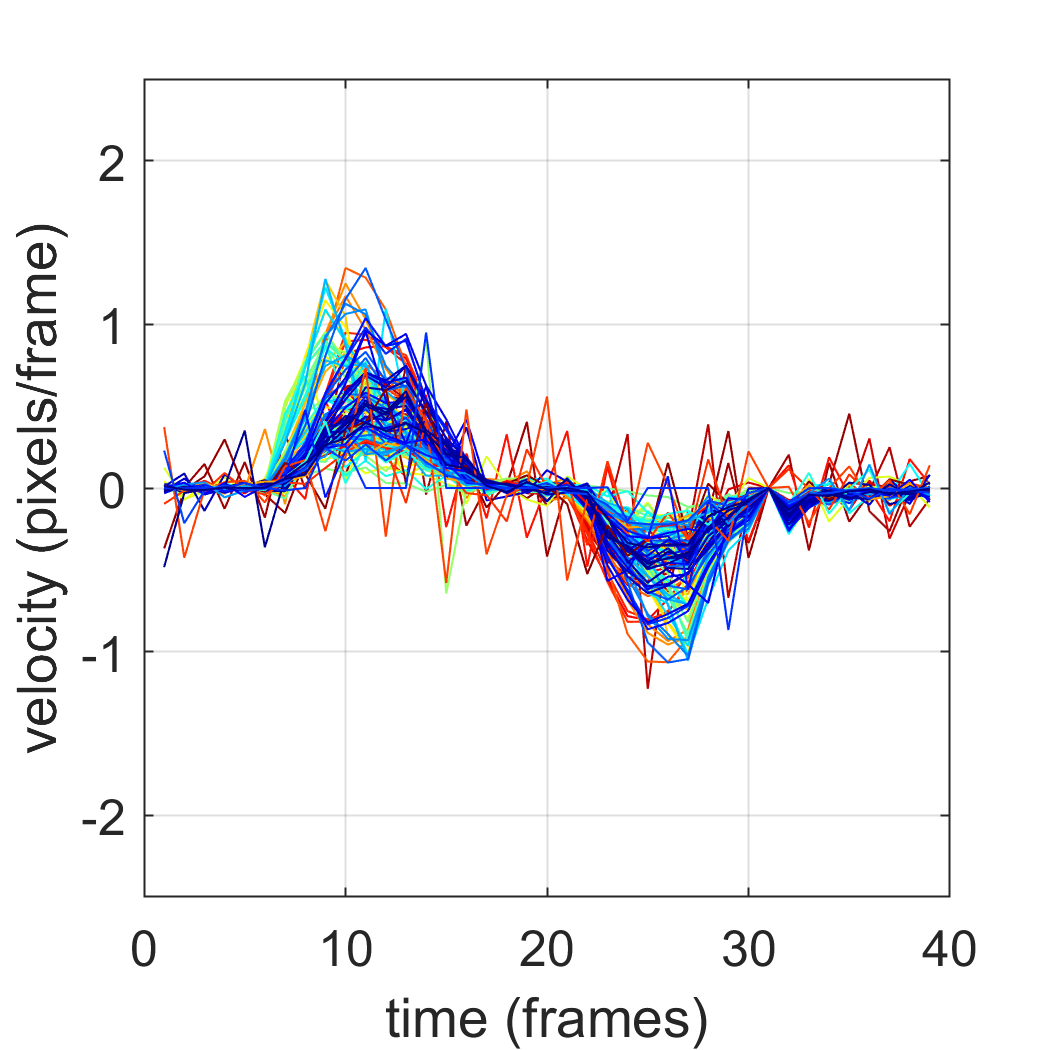}&	
	\includegraphics[width=0.19\textwidth,trim={0 0 15 15},clip]{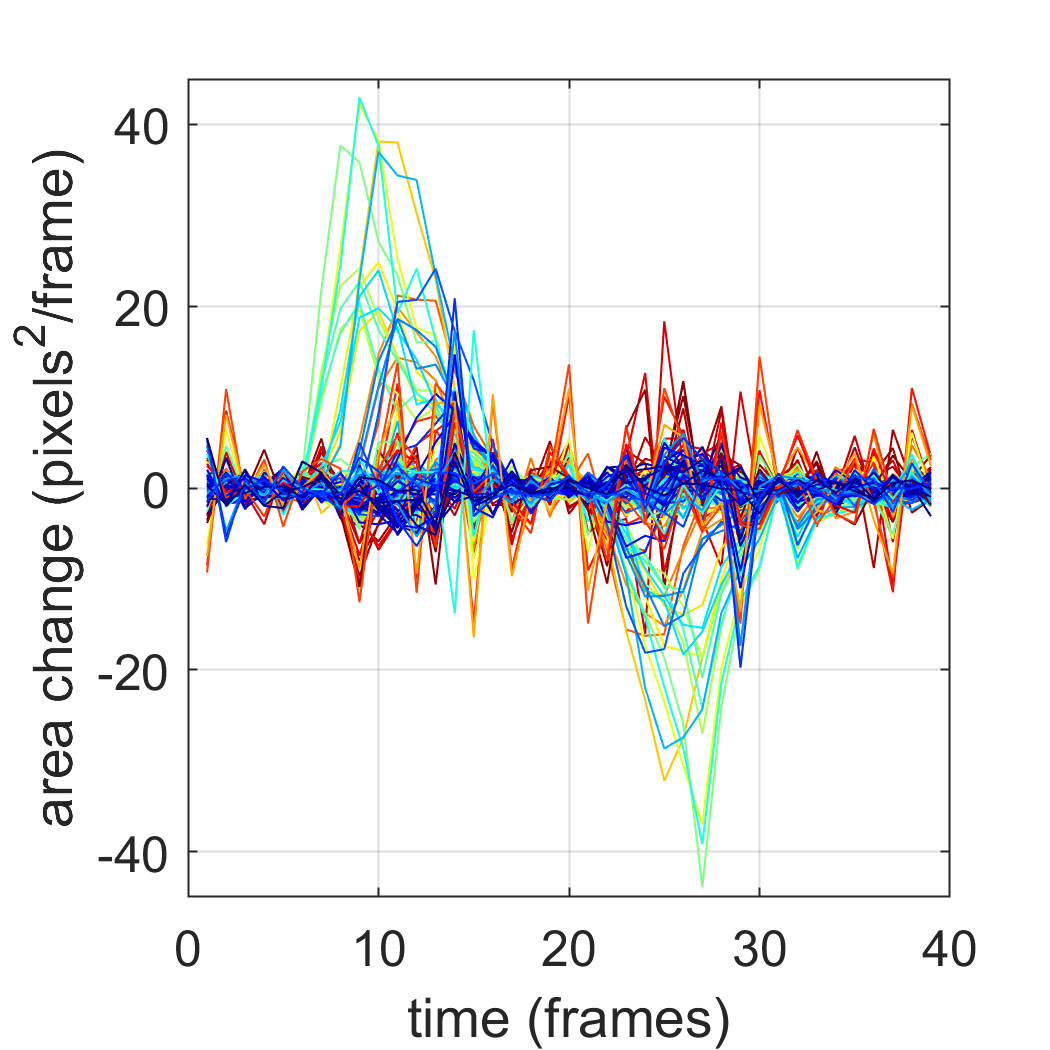} \\
	\end{tabular}\\
	\vspace{0em}
	\caption{Shape sensing with the BRL TacTip. (a) Processing of a tactile image of an edge contact (left) into 2D $(x,y)$-shear strains, from which shear magnitude and Voronoi area change are extracted (red/blue shows positive/negative change; scale in panel b). (b) Biomimetic signal transduction of skin indentation into marker displacement as a shear-strain field. (c) Time series of quantities in panel~a and their derivatives for a contact onto then off an edged stimulus (colour labels marker). This biomimetic representation using shear-strain is highly informative about the contact. }
	\label{fig:6}
\end{figure*}


\section{Tactile Shear Sensing}
\label{sec:4}

\subsection{Tactile sensing of normal strain and shear strain}

A central aspect of tactile sensing with the TacTip is that it measures contact via shear. Skin indentation tilts and moves the nodular pins, causing lateral displacements of markers on the pin tips~(Figure~\ref{fig:6}b, right). The biomimetic structure of stiff nodular pins interdigitating with a soft elastomeric gel causes mechanical transduction of normal-strain on the skin surface into a measurable shear-strain underneath the skin. 

This design also makes the TacTip highly sensitive to shear-strain of the skin surface. The two sources of skin movement, normal and shear strain, produce distinct patterns of marker displacement. Sliding or slip gives a more uniform shear-strain field, whereas normal indentation can produce characteristic dipole or multipole patterns~(Figure~\ref{fig:6}a). In practise, both normal- and shear-strain are usually present, which can introduce subtleties when inferring contact shape information from the TacTip, as discussed later.  

Therefore, the TacTip operates in a distinct manner from the common types of taxel-based artificial tactile sensors that measure normal strain, such as the BioTac~\cite{wettels_biomimetic_2008} and the iCub fingertip~\cite{schmitz_methods_2011}. The TacTip is also distinct from other designs of optical tactile sensor. Reflection-based optical tactile sensors such as the GelSight~\cite{johnson_retrographic_2009} reconstruct normal strain directly from light shading. Some optical tactile sensors also use markers coated underneath the skin~\cite{ferrier_reconstructing_2000} (including recent GelSight versions~\cite{yuan_measurement_2015}), but those measure just the shear strain of the skin conforming to a surface. In contrast, the TacTip markers inform about both the normal and shear strain due to contact. 

Several optical tactile sensors use markers floating in the elastomer underneath the outer skin, such as the GelForce~\cite{kamiyama_evaluation_2004,kamiyama_vision-based_2005} and ChromaTouch~\cite{lin_sensing_2019}. These tactile sensors differ from the TacTip by not having stiff nodular pins connecting the deformation of the skin surface to the markers, but instead rely on the forces transmitted through the soft elastomer. The GelForce uses two layers of coloured spherical markers in a transparent elastomer to infer force vectors within the gel~\cite{kamiyama_evaluation_2004,kamiyama_vision-based_2005}, and the ChromaTouch uses two layers of partially-transparent coloured markers whose movement and colour-matching inform about the sensor deformation~\cite{lin_sensing_2019}. 


All these designs have their pros and cons as artificial tactile sensors. The overall point to emphasise is that the TacTip senses contact differently from other tactile sensors and uses a biomimetic mechanism for this transduction.

\subsection{The shear-sensing hypothesis}

Platkiewicz, Lipson and Hayward have emphasised the importance of internal shear strain for sensing external contact~\cite{platkiewicz_haptic_2016} with their {\em shear-sensing hypothesis}: `we propose that shape-related tactile information is more suitably recovered from shear strain than normal strain'.

Their reasoning is based on the contact mechanics of touch: `the pressure distribution at the surface of a tactile sensor cannot be acquired directly and must be inferred from the [strain] field induced by the touched object in the sensor medium.' In vision, it is well known that edge detection is fundamental to processing shape, with the convolutions of deep neural networks analogous to edge-detecting neurons in visual cortex. For touch, there are two ways to detect edges: indirectly from gradients of the normal-strain field or directly from zero crossings (sign changes) of the internal shear-strain field (Figure~\ref{fig:6}b). Zero crossings are a more robust measure of shape from touch, because they are affected less by distortion from the mechanics of skin and require no computation of signal gradients~\cite{platkiewicz_haptic_2016}, motivating the use of shear strain. 

This reasoning can be grounded in a biomechanical model of skin~\cite{platkiewicz_haptic_2016} that approximates the shear-strain field $\gamma$ at depth $z$ and horizontal position $x$ in a uniform elastomer (Young's modulus $E$) due to a pressure field $p(x)$ at the surface:
\begin{equation}\label{eq1}
\gamma(x,z)\approx \frac{z}{E/3}\frac{{\rm d}}{{\rm d}x}p_\epsilon(x,z),\hspace{1em}p_\epsilon(x,z)=p(x)*\phi_\epsilon(x,z),
\end{equation}
where $p_\epsilon(x,z)$ represents that the pressure field is mechanically blurred by depth (modelled by convolving a Gaussian $\phi_\epsilon(x)$ of width $\epsilon(z)$ that increases with depth). Overall, the shear-strain field follows the gradient of the (blurred) pressure profile and increases linearly with depth from zero at the surface -- {\em i.e.} it has mechanically calculated the signal gradient.

However, the skin of the TacTip is not a uniform elastic material, but has an inner structure of stiff nodular pins interdigitating with soft elastomeric gel. Hence, this model~(\ref{eq1}) is more suited to optical tactile sensors with floating markers such as the GelForce and ChromaTouch~\cite{kamiyama_evaluation_2004,kamiyama_vision-based_2005} 

A biomechanical model of the TacTip and the dermal papillae should be based instead on the levering of the nodular pins caused by the normal skin indentation $\delta(x)$ due to surface pressure. For a pin along the normal to the skin surface (Figure~\ref{fig:6}b), its tip moves horizontally with shear strain:
\begin{equation}\label{eq2}
\gamma(x,z)= z\sin\arctan\left(\frac{{\rm d}}{{\rm d}x}\delta(x)\right)\approx z\frac{{\rm d}}{{\rm d}x}\delta(x),
\end{equation} 
for pins of length $z$ forming the hypotenuse of a right-angled triangle with shear strain $\gamma$ along the (horizontal) adjacent~side. The gradient~$\frac{{\rm d}}{{\rm d}x}\delta(x)$ is tangential to the skin surface and normal to the pins. The shear-strain field is linear in the surface gradient when the small angle approximation $\frac{{\rm d}}{{\rm d}x}\delta(x)\ll 1$ holds; {\em i.e.} it also mechanically calculates the signal gradient.


It is interesting that these two biomechanical models lead to very similar equations: in both, the shear-strain field $\gamma(x,z)$ is proportional to depth $z$ and the gradient of a quantity (pressure or indentation) at the skin surface. Both models amplify the transduction with depth $z$, but the pin structure is beneficial in not being mechanically blurred $\epsilon(z)$ compared to the elastomer. Also,  the shear strain from the stiff pins should be larger than that of an elastic medium. 


Overall, the reasoning behind the shear-sensing hypothesis appears consistent with both the uniform elastomeric model considered originally (Equation~\ref{eq1}) and the pin model introduced here for the TacTip (Equation~\ref{eq2}). The TacTip is unique amongst related tactile sensors by sensing touch purely from shear via a mechanism of stiff pins to transduce skin indentation into shear strain. Therefore, the shear-sensing hypothesis appears to encapsulate the operation of the TacTip as an embodiment of the theory.

\section{Period I (2009-2014): Initial Development of the TacTip}
\label{sec:5.1}

Most of the research and development into soft biomimetic optical tactile sensing has been within the Bristol Robotics Laboratory (BRL) in the U.K. However, the focus has moved from soft to medical to tactile robotics, leading to renewal of the underlying technology and research direction. Meanwhile, the field of tactile robotics has risen in prominence since the TacTip was introduced in 2009 as the aim of the field to enable artificial manipulation has become closer to reality. Recently, the revolution in deep learning~\cite{krizhevsky_imagenet_2017} has raised the prominence of optical tactile sensors because of their compatibility with convolutional neural networks, promising an entirely new level of robot performance approaching human dexterity.

The first period of development (2009-2014) encompassed the original design for an optical tactile sensor based on biologically-inspired encoding~\cite{chorley_development_2009,chorley_tactile_2010}. The name TACTIP was coined~\cite{winstone_tactip_2012,winstone_tactip_2013} as a contraction of `tactile fingertip' after an EU-funded project that supported some of the research. The name has stuck but is now written TacTip similarly to other well-known tactile sensors ({\em e.g.} BioTac or GelSight).

\begin{figure}[t!]
	\centering
	\begin{tabular}[b]{cc}
		\multicolumn{2}{c}{\bf (a) Mechanics of the Tactile Contact Lens} \\
		\includegraphics[width=0.65\columnwidth,trim={0 0 0 5},clip]{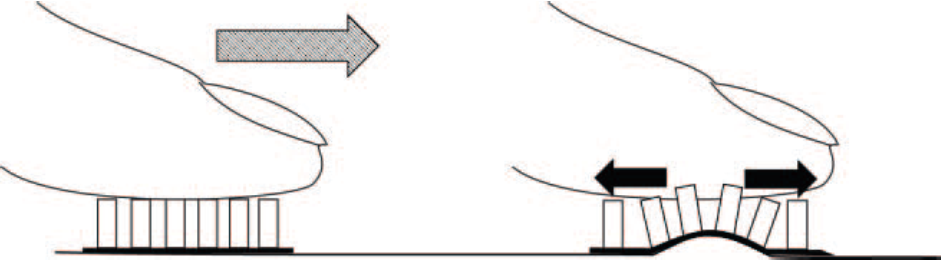} &
		\includegraphics[width=0.28\columnwidth,trim={0 0 0 0},clip]{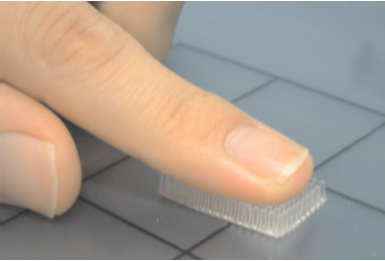} \\
		& \\
		\multicolumn{2}{c}{\bf (b) Mechanics of the Human Fingertip and TacTip} \\
		\includegraphics[width=0.35\columnwidth,trim={100 0 0 180},clip]{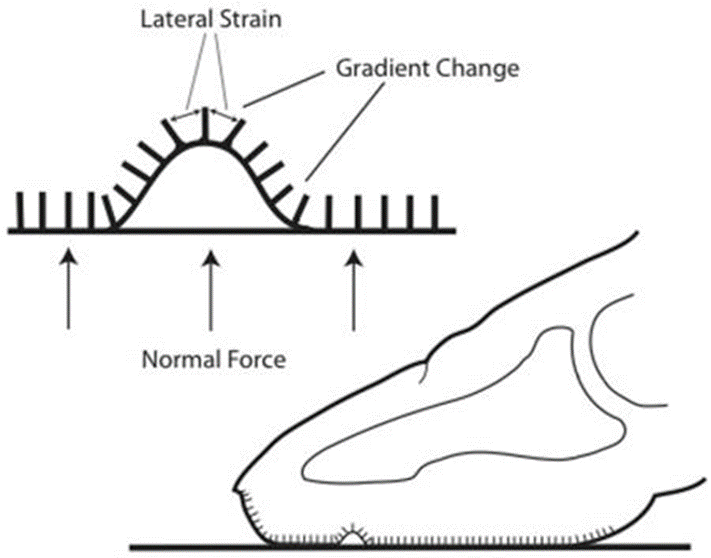} 
		\includegraphics[width=0.3\columnwidth,trim={0 150 100 0},clip]{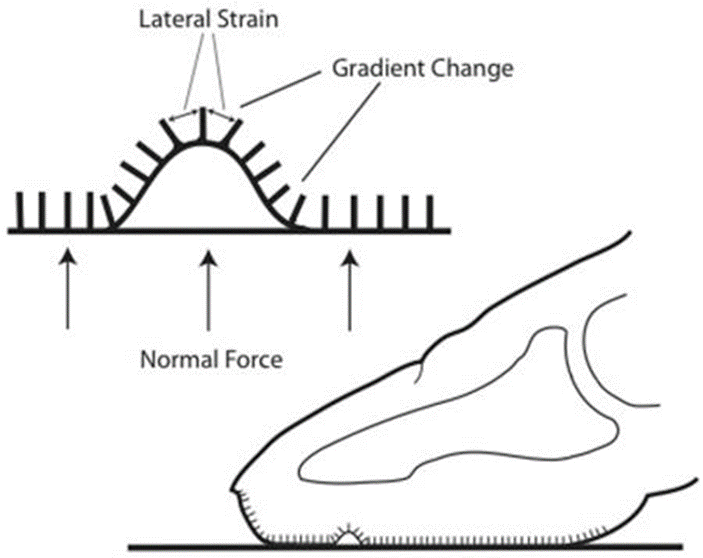} &
		\includegraphics[width=0.28\columnwidth,trim={200 150 205 50},clip]{fig7a} \\
	\end{tabular}
	\caption{Analogous operation of the Tactile Contact Lens and TacTip. Both devices magnify and transduce surface indentation into shear strain, using the levering of an array of pins.}
	\label{fig:5}
	\vspace{-1em}
\end{figure}

The inspiration for the TacTip was the {\em tactile contact lens}~\cite{kikuuwe_tactile_2004,kikuuwe_enhancing_2005} that was developed in a Toyota-funded research laboratory to help humans feel manufacturing defects in automobile production. The tactile contact lens magnifies the feeling of surface deformation via a slippery flexible base-plate attached to an array of rigid pins held against a human fingertip~(Figure~\ref{fig:5}). As the base slides over a small bump or crack, the pins act as levers to amplify the normal indentation in the base into a shear strain that a human can feel~\cite{ando_effect_2016}. Overall, the tactile contact lens magnifies imperceptible surface features into larger shear displacements by about a factor of four, so the contact feature can be discriminated more easily. 

The TacTip uses this same mechanical principle to magnify indentations on a flexible outer skin into lateral movement of markers on the pin tips~\cite{chorley_development_2009}. Both the TacTip and tactile contact lens are biomimetic, because this mechanism for magnifying surface contact into shear strain seems to have evolved in human tactile skin~\cite{cauna_nature_1954}: the human sense of touch relies on the mechanical structure of dermal papillae and epidermal intermediate ridges, as explained in Section~\ref{sec:3}. 

A central question when using any soft optical tactile sensor is how to interpret the tactile image from the camera to infer aspects of the skin deformation (Figure~\ref{fig:6}). Initial studies with the TacTip processed the images into velocity vector fields of marker motion, taking inspiration from the Meissner corpuscles embedded in the dermal papillae which are motion sensitive~\cite{chorley_development_2009,chorley_tactile_2010}. Surface edge features such as the contours of a coin were clearly visible in the marker velocity field, motivating the initial research on biologically-inspired edge encoding. Other early work used image filtering (dilation/erosion) to transform a tactile image into a visualisation of marker density showing the orientation of a contacted edge~\cite{assaf_realtime_2010,assaf_seeing_2014}. Other simple methods to infer contact features were also explored, such as binning the tactile image into a discrete array, which was applied to lump detection and localization in medical haptics~\cite{roke_deformation-based_2011,roke_lump_2012,roke_effects_2013}.


Refinements of the original TacTip design were considered for an EU-funded project that sought to address the need for tactile robot manipulators and probes for industrial and healthcare applications. The tactile tip was miniaturized to a 20\,mm-diameter dome (from 40\,mm) for integration as a fingertip of an anthropomorphic robot hand~\cite{winstone_tactip_2012} and a high frame rate camera used to investigate texture perception~\cite{winstone_tactip_2013}; however, in both cases the camera was separate rather than integrated into the design. These early studies laid the foundation for later advances that used 3D printing to improve the TacTip.


\begin{table*}[t!]
	\centering 
	\begin{tabular}{@{}c@{}c@{}c@{}c@{}} 
		{\bf Sensor} & {\bf Year} & {\bf Design} & {\bf Camera} \\
		\hline 
		{\bf TACTIP}~\cite{chorley_development_2009} & 2009 & \begin{tabular}{c} cylindrical body (40\,mm-dia. spacer) \\[-0.5ex] molded hemispherical soft tip (40\,mm dia., 532 pins) \end{tabular} & LifeCam VX webcam, 480p 30fps, $f$$\approx$50\,mm \\
		Mini TACTIP~\cite{winstone_tactip_2012} & 2012 & \begin{tabular}{c} rigid 3DP mount on digit of Elumotion robotic hand \\[-0.5ex] molded hemispherical soft tip (20\,mm dia., 276 pins) \end{tabular} & Not integrated \\ 
		Open TacTip~\cite{ward-cherrier_tactip_2016} & 2016 & \begin{tabular}{c} rigid 3DP body (161\,mm) modified for modular tip \\[-0.5ex] molded hemispherical soft tip (40\,mm dia., 532 pins) \end{tabular} & LifeCam Cinema webcam, 720p 30fps, $f$$\approx$50\,mm \\
		{\bf TacTip}~\cite{ward-cherrier_tactip_2018} & 2016 & \begin{tabular}{c} rigid 3DP body (85\,mm) and modular tip \\[-0.5ex] hemispherical 3DP soft tip (40\,mm dia., 127 pins) \end{tabular} & LifeCam Cinema board (disassembled camera) \\
		TacTip-M2 \cite{ward-cherrier_tactile_2016} & 2016 & \begin{tabular}{c} integrated as digit of 3DP M2 gripper \\[-0.5ex] rectangular 3DP soft finger (32$\times$102$\times$44\,mm; 80 pins) \end{tabular} & LifeCam Cinema board \\  
		TacCylinder~\cite{winstone_toward_2017} & 2017 & \begin{tabular}{c} 3DP body and soft skin \\[-0.5ex] cylinder (63\,mm dia., 82\,mm length, 180 pins) \end{tabular} & Catadioptric 360$^\circ$ lens; LifeCam Cinema HD \\ 
		TacTip-GR2~\cite{ward-cherrier_model-free_2017} & 2017 & \begin{tabular}{c} integrated as 2 fingertips of 3DP GR2 gripper \\[-0.5ex] 3DP fingertip (40\,mm dia.$\times$ 44\,mm depth, 127 pins)  \end{tabular} & Raspberry Pi Spycam, Fisheye lens, $f$$\approx$20\,mm \\
		TacTip-FP1~\cite{cramphorn_addition_2017}  & 2017 & \begin{tabular}{c} modified tip with `fingerprint' (raised bumps) \\[-0.5ex] hemispherical 3DP soft tip (40\,mm dia., 127 pins) \end{tabular} & LifeCam Cinema board \\  
		TacTip-SYM~\cite{ward-cherrier_exploiting_2017} & 2017 &  \begin{tabular}{c} modified tip with 12-fold rotational symmetry \\[-0.5ex] hemispherical 3DP soft tip (40\,mm dia., 49 pins)  \end{tabular} & LifeCam Cinema board \\  
		TacTip-DM~\cite{pestell_dual-modal_2018} & 2018 & \begin{tabular}{c} customized TacTip for optical flow sensor \\[-0.5ex] small hemispherical 3DP soft tip (28\,mm; 19-61 pins) \end{tabular} & ADNS-3080 dual mode, 30$\times$30\,pix 3fps / 6.4\,kHz \\ 
		MultiTip~\cite{soter_multitip_2018} & 2018 & \begin{tabular}{c} modified tip moulded from thermoactive material \\[-0.5ex] body similar to 2009 TACTIP (40\,mm dia., 127 pins)\end{tabular} & LifeCam VX webcam, 480p\,30fps, $f$$\approx$50\,mm \\ 
		TacWhisker~\cite{lepora_tacwhiskers_2018} & 2018 & \begin{tabular}{c} modified tip with 19 3DP whiskers fitting into pins \\[-0.5ex] also a tendon-actuated version with 2$\times$5 whiskers \end{tabular} & LifeCam Cinema board \\ 
		{\bf TacTip}~\cite{james_slip_2018} & 2018 & same as 2016 version & ELP camera module, 1080p\,30fps\,-\,360p 120fps \\
		TacTip-SMG~\cite{pestell_sense_2019} & 2019 & \begin{tabular}{c} integrated as 3 fingertips of Shadow Modular Grasper \\[-0.5ex] custom 3DP soft fingertips ($\sim$40$\times$40$\times$44\,mm, 100 pins) \end{tabular} &  ELP camera module; wide angle lens $f$$\approx$20\,mm \\ 
		TacTip-Mini~\cite{church_deep_2020} & 2020 & \begin{tabular}{c} modified tip with small area and high pin density \\[-0.5ex] small domed 3DP soft tip (25\,mm dia., 331 pins) \end{tabular} &  ELP camera module \\ 
		TacTip-FP2~\cite{james_slip_2020} & 2020 & \begin{tabular}{c} modified tip with fingerprint to induce incipient slip \\[-0.5ex] ridged 3DP soft tip (40\,mm dia., 44 pins) \end{tabular} &  ELP camera module  \\ 
		TacFoot~\cite{stone_walking_2020} & 2020 & \begin{tabular}{c} integrated as foot of Lynxmotion SQ3U quadruped \\[-0.5ex] small hemispherical 3DP soft tip (28\,mm; 37 pins) \end{tabular} & Hydream USB endoscope, 480p 30fps f$\approx$40\,mm \\
		NeuroTac~\cite{ward-cherrier_neurotac_2020} & 2020 & \begin{tabular}{c} neuromorphic TacTip with event-based output \\[-0.5ex] hemispherical 3DP soft tip (40\,mm dia, 61 pins) \end{tabular} & Inivation DVS240, 240$\times$120pix 12Mevents/s  \\  
		NeuroTac-SoftH~\cite{ward-cherrier_miniaturised_2020} & 2020 & \begin{tabular}{c} integrated as fingertip of Pisa/IIT SoftHand \\[-0.5ex] small 3DP soft fingertip (20$\times$25$\times$30\,mm; 38 pins) \end{tabular}  & Inivation mini-eDVS, 128$\times$128pix 0.6Mevents/s \\  
		TacTip-O~\cite{james_tactile_2020} & 2020 & \begin{tabular}{c} integrated as 3 fingertips of 3DP Model-O hand \\[-0.5ex] small 3DP soft fingertips (20$\times$40$\times$35\,mm, 30 pins) \end{tabular} & \begin{tabular}{c} JeVois Vision module \\[-0.5ex] 1280$\times$1024pix 15fps to 276$\times$144pix 120fps \end{tabular}   \\ 
		TacTip-SoftH~\cite{lepora_towards_2021} & 2021 & \begin{tabular}{c} integrated as fingertip of Pisa/IIT SoftHand \\[-0.5ex] small 3DP soft fingertip (12$\times$19$\times$17\,mm; 35 pins) \end{tabular} &  Misumi Model SYD, 1080p 60fps f$\approx$10\,mm \\				
		{\bf TacTip} & 2021 & \begin{tabular}{c} rigid 3DP body (45\,mm) and modular tip \\[-0.5ex] hemispherical 3DP soft tip (40\,mm dia., 330 pins) \end{tabular}  & ELP camera module; wide angle lens $f$$\approx$10\,mm  \\
	\end{tabular}
	\vspace{0em}
	\caption{TacTip family of soft biomimetic optical tactile sensors. Key: 3D-printed (3DP), 640$\times$480 pix (480p), 1280$\times$720 pix (720p), 1920$\times$1080 pix (1080p), frames per second (fps), focal~length~($f$). LifeCam CCD webcams by Microsoft, ELP (Ailipu Technology) USB CMOS camera modules, Dynamic Vision Sensors (DVS) are event-based cameras.}
	\label{table:2} 
	\centering 
	\vspace{1em}
	\begin{tabular}{@{}cccc@{}} 
		{\bf Hand} & {\bf DoA} & {\bf Fingers} & {\bf Examined capabilities} \\
		\hline
		Tactile Model M2 (T-M2) & 1 & 2 (1 tactile) & precise rolling manipulation~\cite{ward-cherrier_tactile_2016,ward-cherrier_tactip_2018} \\
		Tactile Model GR2 (T-GR2) & 2 & 2 tactile  & precise rolling manipulation~\cite{ward-cherrier_model-free_2017,ward-cherrier_tactip_2018} \\	
		Tactile Model O (T-MO) & 4 & 3 tactile & \begin{tabular}{c} retains grasping capabilities; tactile object recognition; grasp success prediction~\cite{james_tactile_2020}\\[-0.5ex] slip detection \& correction; light grasping on first attempt~\cite{james_slip_2020} \end{tabular} \\
		Tactile Modular Grasper (T-MG) & 9 & 3 tactile & \begin{tabular}{c} manipulation to desired grasp~\cite{pestell_sense_2019} \end{tabular} \\	
		Tactile SoftHand (T-SoftH) & 1 & 5 (1 tactile) & \begin{tabular}{c} sensorimotor control of touch~\cite{lepora_towards_2021}\\[-0.5ex] tactile object recognition; grasp perturbation recognition~\cite{ward-cherrier_miniaturised_2020} \end{tabular} \\	
	\end{tabular}
	\vspace{0em}
	\caption{Integration of the TacTip into robot hands varying in degrees of actuation (DoA) and numbers of fingertips.} 
	\label{table:3} 
\end{table*}	

\begin{figure*}[t!]
	\centering
	\begin{tabular}[b]{c}
		{\bf SoftBOT Sensors: Soft Biomimetic Optical Tactile Sensors} \\ 	\vspace{-1.5em}
		\includegraphics[width=0.95\textwidth,trim={0 -15 0 0},clip]{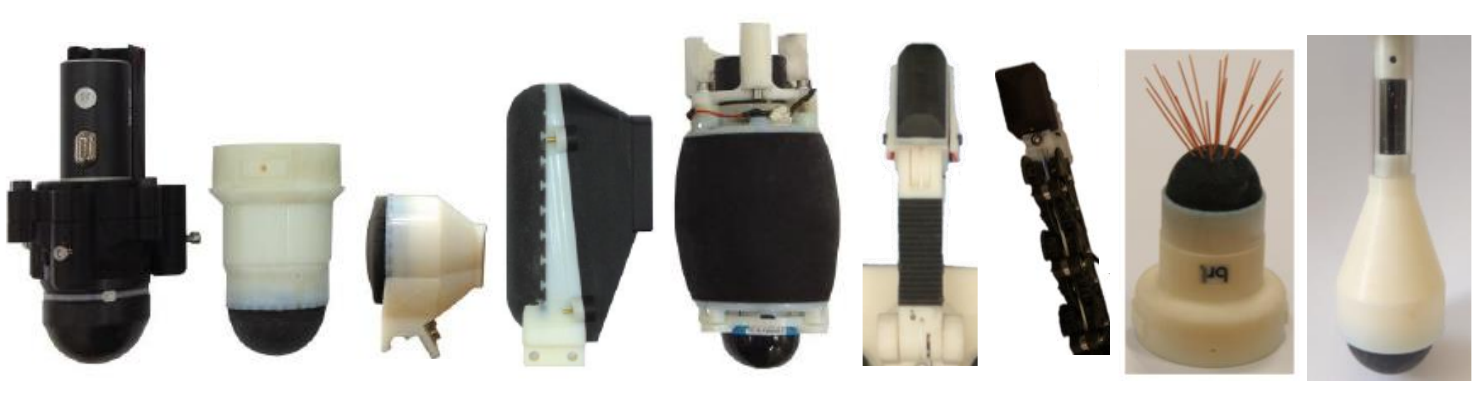}\\
		{\bf SoftBOT Hands: Soft Biomimetic Optical Tactile Hands} \\   \vspace{-1.5em}
		\includegraphics[width=0.95\textwidth,trim={0 -15 0 -5},clip]{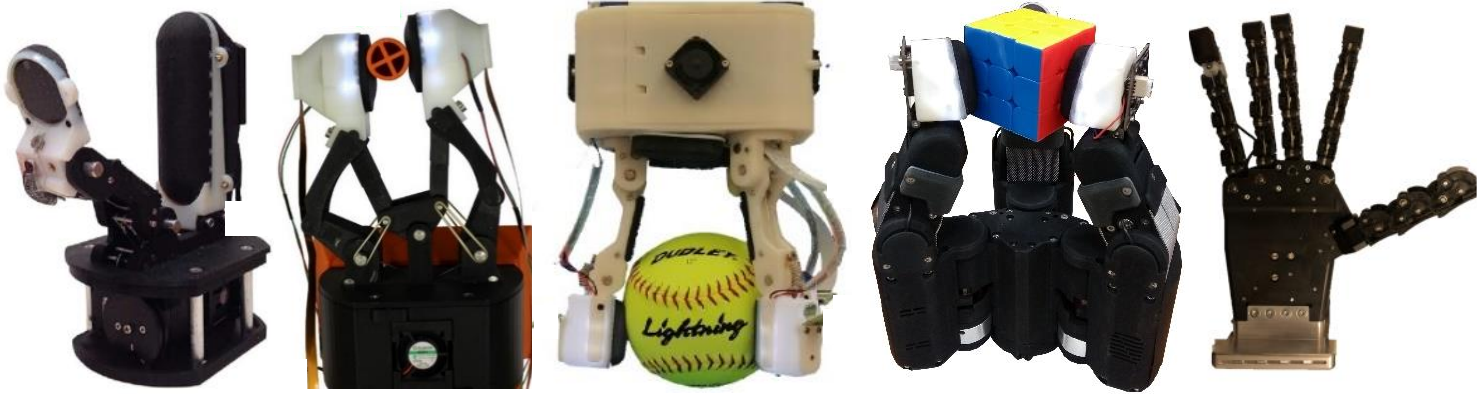}\\  
		{\bf SoftBOT Systems: Soft Biomimetic Optical Tactile Systems} \\   
		\includegraphics[width=0.95\textwidth,trim={0 0 0 -5},clip]{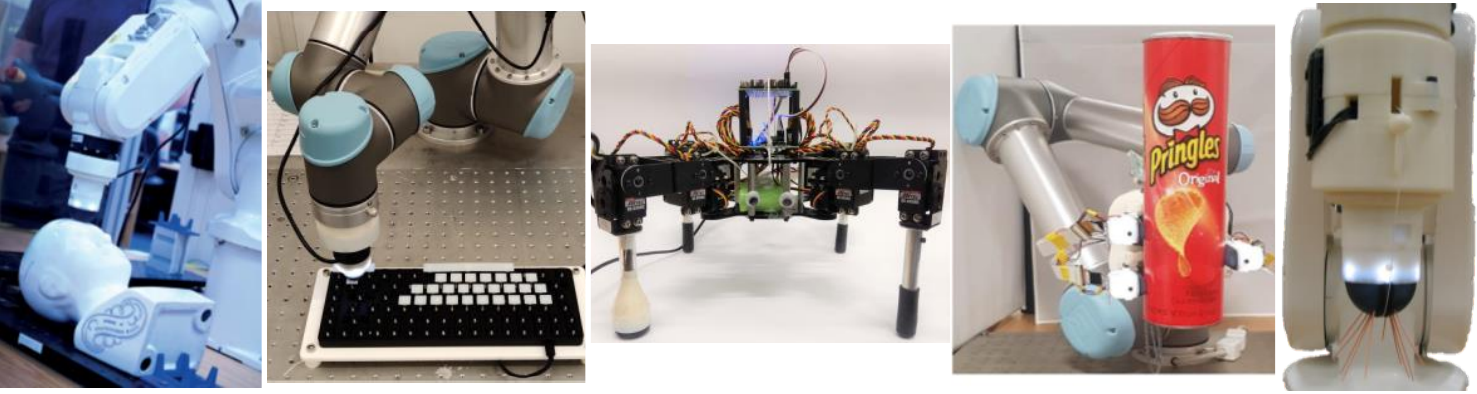}\\
	\end{tabular}
	\caption{SoftBOTS from the TacTip family. SoftBOT sensors (top row): original 2009 TACTIP~\cite{chorley_development_2009}, 3D-printed TacTip~\cite{ward-cherrier_tactip_2018}, TacTip-GR2~\cite{ward-cherrier_model-free_2017}, TacTip-M2~\cite{ward-cherrier_tactile_2016}, TacCylinder~\cite{winstone_toward_2017}, TacTip-O~\cite{james_tactile_2020}, TacTip-SoftHand~\cite{lepora_towards_2021}, TacFoot~\cite{stone_walking_2020} and TacWhisker~\cite{lepora_tacwhiskers_2018}. SoftBOT hands (middle row): tactile Model-M2~\cite{ward-cherrier_tactile_2016}, tactile Model-GR2~\cite{ward-cherrier_model-free_2017}, tactile Model-O~\cite{james_tactile_2020}, tactile Shadow Modular Grasper~\cite{pestell_sense_2019} and tactile SoftHand~\cite{lepora_towards_2021}. SoftBOT systems (bottom row): 3D-printed TacTip on ABB robot arm for tactile servo control~\cite{lepora_exploratory_2017,lepora_pixels_2019}, TacTip on UR5 arm for reinforcement learning~\cite{church_deep_2020}, TacFoot on walking robot~\cite{stone_walking_2020}, tactile Model-O on UR5 robot arm for grasping and slip detection~\cite{james_tactile_2020,james_slip_2020} and tacWhisker mounted on ABB robot arm~\cite{lepora_tacwhiskers_2018}.}
	\label{fig:4}
\end{figure*}


\section{Period II (2015-): The TacTip Family}
\label{sec:5.2}

The second period of the TacTip development (2015-present) introduced some common approaches that allowed research on soft biomimetic optical tactile sensing at BRL to build into a coherent body of work. Multi-material 3D printing was adopted for fabrication alongside modular design principles~\cite{ward-cherrier_tactip_2018}, which encouraged diversification into a family of tactile sensors, hands and robotic systems (Figure~\ref{fig:4}). A common method based on tracking the markers was used for tactile image processing~\cite{lepora_superresolution_2015,lepora_biomimetic_2016}, so the data could be treated as multi-dimensional time series for application of standard machine learning methods. These developments enabled the TacTip to be integrated into complete robotic systems that perceive, explore and manipulate their environments. 

\subsection{3D-printed TacTip and integration into robot hands}
\label{sec:5.2.1}

The adoption of multi-material 3D printing for TacTip manufacture~\cite{ward-cherrier_tactip_2018} was key to developing a family of rapidly-prototyped tactile probes, grippers and manipulators~(Figure~\ref{fig:4}). The soft biomimetic optical tactile sensor was redesigned (2016/2018 TacTip; Table~\ref{table:2}) to have a compact modular base housing the circuit board from a web-camera; likewise, the tip was 3D-printed in one piece combining a flexible skin joined to papillae tipped with rigid white markers~\cite{ward-cherrier_tactip_2018}. This redesigned TacTip became the standard device used in BRL for tactile sensing research. The 3D-printed skin is robust to laboratory testing, with the tip only needing replacing after human error in operating industrial robot arms (even then, small tears have been repaired).

3D-printing also enabled the sensor to be customised for diverse applications, from creating a tactile sensing foot for walking robots~\cite{stone_walking_2020} to mimicking rodent tactile whiskers~\cite{lepora_tacwhiskers_2018} (Figure~\ref{fig:4}, bottom row). A more biomimetic version of the TacTip skin with raised bumps over the pins (like a fingerprint) and increased dermal-epidermal stiffness contrast (rigid cores to the pins) improved the spatial acuity of the tactile sensor~\cite{cramphorn_addition_2017}. Further progression to a ringed biomimetic fingerprint helped induce and detect incipient slip, by encouraging the outer contact region to move before global slip occurs, giving sufficient time to react before losing the grasp~\cite{james_biomimetic_2020}. 

These advances in soft tactile sensors complement the rapid progress in 3D-printed robot hands, examplified by the Yale OpenHand Project: a library of low-cost 3D-printed underactuated hand designs~\cite{ma_yale_2017}. These tendon-driven, 2-4 fingered compliant hands have an underactuated adaptability that passively conforms their grasping to a wide range of object geometries using only open-loop control. The most well-known is the 3-fingered Model O, based on the iHY (iRobot-Harvard-Yale) Hand~\cite{odhner_compliant_2014} from the DARPA Robotic Manipulation-Hardware (ARM-H) program, which has been commercialised as the Reflex Hand (RightHand robotics). The hand combines an underlying capability for underactuated grasping~\cite{dollar_highly_2010} with sufficient degrees-of-actuation for the manipulation tasks set by the ARM-H program~\cite{odhner_compliant_2014} (5 for the iHY Hand, reduced to 4 for the Model O). 

Integration of the TacTip with the OpenHands (Table~\ref{table:3}) began with the Model M2~\cite{ma_m2_2016}, a relatively simple gripper with just one movable finger and a large immobile thumb that was replaced with the TacTip-M2~\cite{ward-cherrier_tactile_2016}: an elongated tactile sensor with a rectangular sensing surface~(Figure~\ref{fig:4}; middle row). Next, the two-fingered GR2 gripper~\cite{rojas_gr2_2016} was integrated ~\cite{ward-cherrier_model-free_2017}, replacing each fingertip with the TacTip-GR2: a compact version of the TacTip body using a fisheye lens and small camera with a standard-sized tip. Recently, the OpenHand tactile integration has culminated with the Tactile Model O (T-MO)~\cite{james_tactile_2020}: the three fingertips were each replaced with a miniaturized TacTip, offering a low-cost 3D-printed dexterous robot hand with multipurpose soft optical tactile sensing.

A key question when considering the integration of tactile sensing into robotic systems is: what new capabilities are given by the sense of touch? For the 2-fingered tactile OpenHands (T-M2 and T-GR2), the integrated tactile sense enabled precise in-hand manipulation for unknown held objects~\cite{ward-cherrier_tactile_2016,ward-cherrier_model-free_2017}. The fingers could roll an object to a desired location or along a trajectory over the tactile fingertip to millimetre accuracy~\cite{ward-cherrier_tactip_2018}. The 3-fingered T-MO's capabilities are based around the hand's high capability at grasping. Applying supervised deep learning to the tactile images of grasped objects gave accurate object classification (93\%, 26 objects) and grasp success prediction for lifting (95\%, same objects)~\cite{james_tactile_2020}. Furthermore, the application of slip-detection methods enabled the hand to quickly re-grasp slipping objects (11 objects; 6 novel, 1 compliant) before being dropped~\cite{james_slip_2020} (see Section~\ref{sec:5.2.2}). The tactile hand could also prevent an object being dropped when weight was added, {\em e.g.} rice poured into a held tube, and lift an object on the first attempt with minimal grasp force~\cite{james_slip_2020}.

Two other state-of-the-art hands have also been integrated with TacTips (Table~\ref{table:3}): the Shadow Smart Modular Grasper and the Pisa/IIT SoftHand. The Smart Modular Grasper is a fully-actuated dexterous 3-fingered hand that was combined with a compact TacTip using a wide-angle camera lens and customized body and tip from the 2018 TacTip, which has been applied to stabilizing grasps by controlling the fingertip contacts~\cite{pestell_sense_2019,psomopoulou_robust_nodate}. The SoftHand is an underactuated anthropomorphic robot hand designed around the principle of adaptive synergies from human hand movements~\cite{catalano_adaptive_2014}. This 5-fingered hand has been integrated with a miniaturized TacTip-SoftH of similar size to a human fingertip~\cite{lepora_towards_2021}, a key milestone in this family of tactile hands (Figure~\ref{fig:4}; middle-row). A slightly-larger TacTip with an event-based camera (NeuroTac) has also been integrated~\cite{ward-cherrier_neurotac_2020}. Both versions of the tactile SoftHand open new possibilities for the sensorimotor control of anthropomorphic robot hands with biomimetic touch. 

\begin{figure*}[t!]
	\centering
	\renewcommand{\tabcolsep}{2pt}
	\vspace{2em}
	\begin{tabular}[b]{@{}ccc@{}}
		{\bf (a) Pose-based Servo Control via} & {\bf (b) Goal-based pushing via} & {\bf (c) Tactile Deep RL:} \\
		{\bf Tactile Deep Learning}~\cite{lepora_optimal_2020,lepora_pose-based_2020} & {\bf tactile \& proprioceptive feedback}~\cite{lloyd_goal-driven_2020} & {\bf Learning to type Braille}~\cite{church_deep_2020}\\
		\includegraphics[width=0.32\textwidth,trim={0 0 0 0},clip]{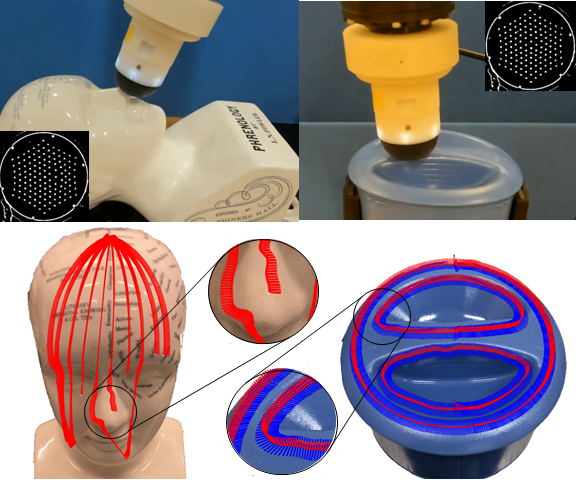}&
		\includegraphics[width=0.32\textwidth,trim={0 0 0 0},clip]{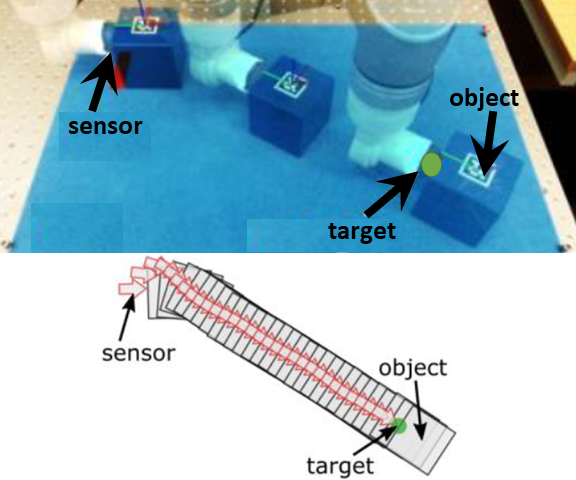}&
		\includegraphics[width=0.32\textwidth,trim={0 0 0 0},clip]{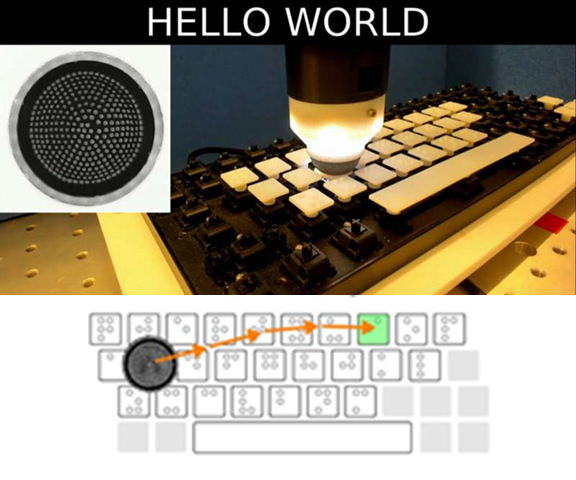}\\
	    {\bf (d) Object \& grasp-success prediction}&{\bf (e) In-hand tactile manipulation for}&{\bf (f) Fine control of contact with} \\
		{\bf with an underactuated tactile hand}~\cite{james_tactile_2020}  &{\bf fully-actuated stable grasping}~\cite{psomopoulou_robust_nodate} &{\bf an anthropomorphic tactile hand}~\cite{lepora_towards_2021} \\
		\includegraphics[width=0.32\textwidth,trim={0 0 0 0},clip]{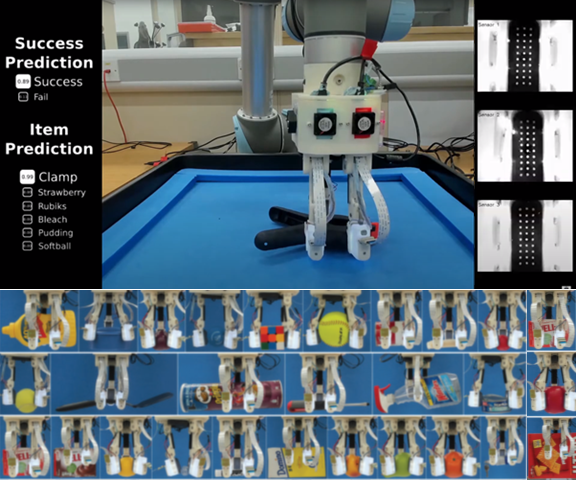} &
		\includegraphics[width=0.32\textwidth,trim={0 0 0 0},clip]{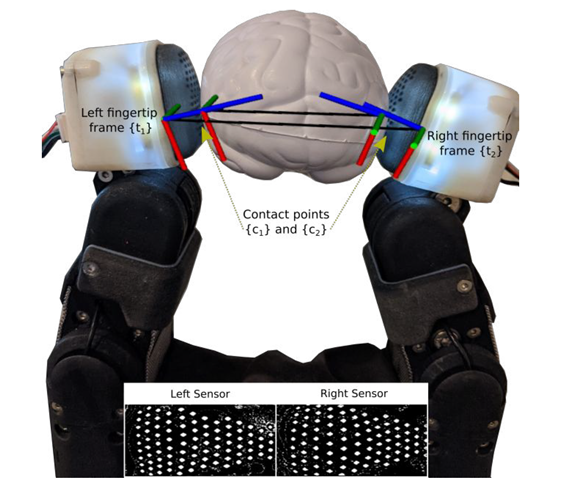} &
        \includegraphics[width=0.32\textwidth,trim={0 0 0 0},clip]{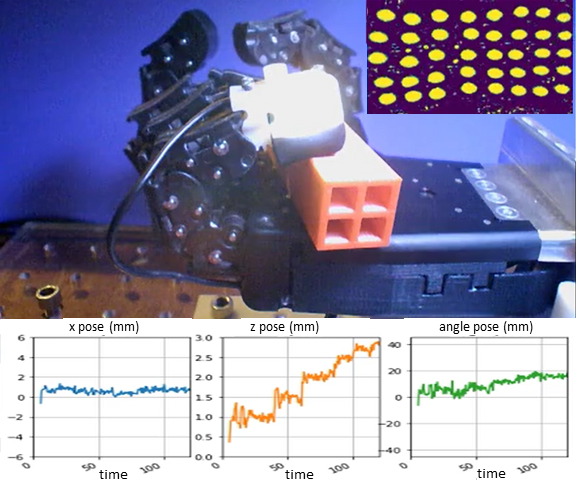} \\
	\end{tabular}
	\caption{Capabilities of deep learning with the BRL TacTip. The top row (a-c) shows examples of using a single tactile sensor mounted on a robot arm. The bottom row shows examples with tactile robot hands: (d) the Tactile Model O, (e) Tactile Modular Grasper and (f) the Pisa/IIT SoftHand. For the diverse examples shown in this figure, the use of convolutional neural networks was critical for the tactile capability to be reached.}
	\label{fig:8}
\end{figure*}
\vspace{1em}
\begin{table*}[h!]
	\centering 
	\begin{tabular}{@{}ccccc@{}} 
		{\bf Capability} & {\bf Year} & {\bf Accuracy} & {\bf Robot} & {\bf Method} \\
		\hline
		1D localization & 2015 & 0.7\,mm & TacTip on ABB arm & probabilistic classifier on pin displacements~\cite{lepora_superresolution_2015,lepora_biomimetic_2016} \\
		Rolling manipulation & 2016 & $\sim$1\,mm & TacTip on ABB arm & probabilistic classifier on pin displacements~\cite{cramphorn_tactile_2016} \\
		Gap measurement & 2016 & 0.5\,mm & TacTip on ABB arm & probabilistic classifier on pin displacements~\cite{lepora_tactile_2016} \\
		Rolling manipulation & 2016 & $\sim$1\,mm & Tactile Model M2 hand & probabilistic classifier on pin displacements~\cite{ward-cherrier_tactile_2016} \\
		2D contour following (tapping) & 2017 & $\sim$1\,mm, $\sim$10\,deg & TacTip on ABB arm & probabilistic classifier on pin displacements~\cite{lepora_exploratory_2017} \\
		Rolling manipulation & 2017 & $\sim$1\,mm & Tactile Model GR2 hand & probabilistic classifier on pin displacements~\cite{ward-cherrier_model-free_2017} \\
		1D localization; curvature estimation & 2017 & 0.1\,mm; $\sim$1\,mm & TacTip-FP1 on ABB arm & probabilistic classifier on pin displacements~\cite{cramphorn_addition_2017} \\
		Slip detection & 2018 & $>$95\% & TacTip on UR5 arm &  SVM classifier on pin velocities~\cite{james_slip_2018} \\
		2D contour following (sliding) & 2019 & $\sim$1\,mm, $\sim$10\,deg & TacTip on ABB arm & PCA \& GP regressor on pin displacements~\cite{aquilina_principal_2018,aquilina_shear-invariant_2019} \\
		2D contour following (sliding) & 2019 & $\sim$1\,mm, $\sim$10\,deg & TacTip on ABB arm & ConvNet regressor on 128$\times$128 tactile image~\cite{lepora_pixels_2019} \\
		Grasp control (centre of contact) & 2019 & $\sim$1\,deg & Tactile Modular Grasper & polynomial regressor on pin displacements~\cite{pestell_sense_2019}\\
		1D localization; Normal force & 2019 & $\sim$0.5\,mm, $\sim$0.2\,N & TacTip on ABB arm & ConvNet autoencoder on ]32$\times$32 tactile image~\cite{polic_convolutional_2019} \\
		\begin{tabular}{c}Object recognition \\[-0.5ex] Grasp success prediction\hspace{-2em}\end{tabular} & 2020 &  \begin{tabular}{c} $>$95\%\\[-0.5ex] $>$90\% \end{tabular}  & Tactile Model O hand & ConvNet classifier on 60$\times120$ tactile image~\cite{james_tactile_2020} \\
		Slip detection & 2020 & $>$90\% & Tactile Model O hand &  SVM or logReg classifier on pin velocities~\cite{james_slip_2018}\\
		Ultrasound detection & 2020 & $\lesssim$10\,micron & TacTip on ABB arm & GP regressor on pin displacements~\cite{alakhawand_sensing_2020} \\
		\begin{tabular}{c} 3D surface localization \\[-0.5ex] (depth, roll/pitch) \end{tabular} & 2020 & \begin{tabular}{c}0.1\,mm \\[-0.5ex] 0.3\,deg\end{tabular}  & TacTip on ABB arm & ConvNet regressor on 128$\times$128 tactile image~\cite{lepora_optimal_2020,lepora_pose-based_2020} \\
		\begin{tabular}{c} 3D edge localization \\[-0.5ex] (horizontal, depth, roll/pitch, yaw) \end{tabular} & 2020 & \begin{tabular}{c}0.3\,mm, 0.2\,mm \\[-0.5ex] 1-2\,deg, 4\,deg\end{tabular} & TacTip on ABB arm & ConvNet regressor on 128$\times$128 tactile image~\cite{lepora_optimal_2020,lepora_pose-based_2020} \\
		Grasp control (stable pinch) & 2021 & 0.1\,mm, 0.4\,deg & Tactile Modular Grasper & servo control \& ConvNet on 128$\times$128 tactile image~\cite{psomopoulou_robust_nodate}\\
		\begin{tabular}{c}3D surface following (sliding) \\[-0.5ex] 3D contour following (sliding) \end{tabular} & 2021 & \begin{tabular}{c}0.1-0.5\,mm\\[-0.5ex] 1-5\,deg\end{tabular} & TacTip on ABB arm & \begin{tabular}{c}pose-based tactile servo control \&\\[-0.5ex] ConvNet on 128$\times$128 tactile image~\cite{lepora_optimal_2020,lepora_pose-based_2020} \end{tabular}  \\
		2D pushing (tapping motion) & 2021 & $\sim$1\,mm & TacTip on UR5 arm &  servo control \& ConvNet on 128$\times$128 tactile image~\cite{lloyd_goal-driven_2020} \\		
	\end{tabular}
	\vspace{0em}
	\caption{TacTip capabilities in order of development. Key: Support Vector Machine (SVM), Principal Component Analysis (PCA), Gaussian Process (GP), Convolutional Neural Network (ConvNet), Logistic Regression (LogReg). The ABB arm is an IRB120 6-DoF industrial robot; the UR5 arm is a Universal Robotics 6-DoF arm. Tactile Hands are described in Table~\ref{table:3}.} 
	\label{table:4} 
\end{table*}

\vspace{-1.5em}
\subsection{Progress in tactile capabilities}
\label{sec:5.2.2}

The purpose of an artificial sense of touch is to impart new dexterous capabilities to robotic systems that physically interact with their surroundings. These capabilities are built on sensorimotor perception and control: perception to process the tactile sensations for inferring states of the environment relative to the sensor, such as the location or shape of an object feature; and control to change that state, such as to slide a fingertip delicately over a feature or manipulate an object in a desired manner. A range of tactile perception and control methods have been developed for the TacTip family of soft biomimetic optical tactile sensors and hands~(Table~\ref{table:4}). 


Over the period 2015-19, most of the methods used time series of marker deflections from rest, as they gave an efficient representation of the tactile image that has a biomimetic analogue with mechanoreceptor activity~\cite{lepora_superresolution_2015,lepora_biomimetic_2016,cramphorn_addition_2017,ward-cherrier_tactip_2018} (see Section~\ref{sec:3}). The pin displacements are also easily visualised to help interpret the tactile sensing (Figure~\ref{fig:6}). 


Over the same period, a perception method based on a histogram likelihood model over the marker displacements was used~\cite{lepora_superresolution_2015}. For control, the position of an object feature was predicted, such as a cylinder position or edge orientation~\cite{lepora_biomimetic_2016}, then applied to rolling manipulation~\cite{cramphorn_tactile_2016,ward-cherrier_tactile_2016,ward-cherrier_model-free_2017} or tapping around 2D contours~\cite{lepora_exploratory_2017,ward-cherrier_exploiting_2017,pestell_dual-modal_2018}. However, those methods did not extend well from discrete to continuous and dynamic environments. Consequently, techniques interpolating over continuously-labelled data were tried, such as Gaussian process regression to control sliding motion around 2D contours~\cite{aquilina_principal_2018,aquilina_shear-invariant_2019} and polynomial regression to smoothly control a multi-fingered grasp~\cite{pestell_sense_2019}. The large amount of training data can be an issue in some circumstances, and so online learning using Gaussian Process latent variable models has also been explored  \cite{stone_learning_2020,stone_walking_2020}. Methods for dynamic environments were also developed, such as accurately detecting and correcting for object slippage~\cite{james_slip_2018,james_slip_2020,james_biomimetic_2020} using a support-vector machine. 


The progress in machine learning methods has steadily improved the TacTip accuracy. Initial performance of $\sim$1\,mm for 1D localization~\cite{lepora_superresolution_2015,lepora_biomimetic_2016,lepora_exploratory_2017} or rolling manipulation~\cite{cramphorn_tactile_2016,ward-cherrier_tactile_2016,ward-cherrier_model-free_2017} has improved by an order-of-magnitude to 0.1-0.5\,mm~\cite{lepora_optimal_2020}. Therefore, while the first results demonstrated hyperacuity finer than the pin spacing~\cite{lepora_tactile_2015,lepora_superresolution_2015}, the latest results demonstrate sensitivity at the level of pixels on the tactile image. Thus far, the best sensitivity with a TacTip has been to detect $\lesssim$10\,microns indentation from an ultrasound haptic display, using signal averaging over the pin deflections from multiple tactile measurements to reduce noise~\cite{alakhawand_sensing_2020}. 




\section{Period III (2019-): Deep learning with the TacTip}
\label{sec:5.3}

Recently, the capabilities of the TacTip have undergone a step-change with the adoption of deep learning over tactile images. This has enabled, for the first time, tactile interaction in real time with complex objects in 3D (Figure~\ref{fig:8}). Prior to deep learning, the capabilities were limited to basic demonstrations such as rolling objects in 1D or exploring around flat shapes in 2D (Table~\ref{table:4}). The new dexterous capabilities are a step closer to those we possess as humans.

The main benefit of deep learning is to predict quantities of interest directly from the tactile images while being insensitive to unknown variations that might otherwise interfere with the predictions. Consequently, the most advanced robot dexterity with the TacTip (Table~\ref{table:4}) has mainly used convolutional neural networks (ConvNets), following their earlier success with optical tactile sensors based on the GelSight~\cite{yuan_shape-independent_2017,calandra_feeling_2017}. In principle, the marker deflections (Section~\ref{sec:5.2.2}) could be used as a lower-dimensional input to a neural network, although in practise it has been simpler to avoid the additional image processing step by using the entire tactile image as input.    

New tactile capabilities developed with the TacTip include: (a) pose-based servo control, where a tactile fingertip mounted on a robot arm slides delicately over unknown complex 3D objects (Figure~\ref{fig:8}a; \cite{lepora_pixels_2019,lepora_optimal_2020,lepora_pose-based_2020}); (b) pushing manipulation of unknown objects using only tactile sensing and proprioceptive knowledge of where the sensor is positioned relative to a goal location (Figure~\ref{fig:8}b; \cite{lloyd_goal-driven_2020}); (c) acquiring the novel skill of single-fingered typing on a braille keyboard, learning the identity of keys and how to navigate the keyboard from touch (Figure~\ref{fig:8}c; \cite{church_deep_2020}); (d) item recognition and grasp-success prediction upon using the tactile sense of the three fingertips of the Tactile Model-O (T-MO) hand (Figure~\ref{fig:8}d; \cite{james_tactile_2020}); (e)~in-hand manipulation of unknown objects to a stable grasp configuration with a fully-actuated tactile Shadow Modular Grasper (Figure~\ref{fig:8}e; \cite{psomopoulou_robust_nodate});  (f) fine control of contact onto unknown objects placed in-hand using an anthropomorphic tactile hand based on the Pisa/IIT SoftHand (Figure~\ref{fig:8}f; \cite{lepora_towards_2021}). Humans can do all these tasks with our sense of touch, which deep learning has enabled tactile robots to likewise perform.


Why are ConvNets so useful when applied to tactile images from a soft biomimetic optical tactile sensor? Considering all the tasks in Figure~\ref{fig:8}, several reasons emerge:\\
\noindent {\em (i)~Ease of deployment} -- in all tasks, the trained neural network was applied directly with only basic pre-processing steps such as cropping, thresholding or concatenating tactile images. This simplifies the algorithmic pipeline and leaves less room for software or other issues to emerge.\\
\noindent {\em (ii)~Robustness} -- once trained, the predictions were relatively unaffected by uncontrolled variations that could cause issues, such as changes in ambient and internal lighting, wear of the skin, visible dust inside the sensor, and even after damage such as a (glued) tear in the TacTip skin after accidental crushing.\\
\noindent {\em (iii)~Ease of scalability} -- the methods seem to extend straightforwardly from simple test scenarios to more complex realistic situations; for example, 2D servo control using a ConvNet with two pose variables extended relatively easily to 3D control with more pose variables~\cite{lepora_pixels_2019,lepora_optimal_2020}.\\
\noindent {\em (iv) Generalization} -- the predictions seemed to be accurate even after large variations in the stimulus or task conditions; for example, ConvNets trained on simple stimuli, such as the pose of a planar surface or straight edge, maintained predictive performance on curved complex stimuli, enabling novel objects to be explored or manipulated~\cite{lepora_pose-based_2020,lloyd_goal-driven_2020}.



Going forward, a key question for deep learning with optical tactile sensing will be on the appropriate learning algorithm to acquire tactile skills. All but one of the tasks in Figure~\ref{fig:8} trained the neural network model with supervised learning. However, a complementary approach is to use tactile deep reinforcement learning (RL) with a reward signal that indicates the desired interactions with a physical environment. The third task (Figure~\ref{fig:8}d) successfully applied this approach to learning to type with a braille keyboard~\cite{church_deep_2020}, leading after many hours of training to a policy network that could efficiently navigate the keyboard to press a desired key. In principle, all other tasks could also be learnt using deep RL, but the long training times are impractical on physical robots. Thus, our expectation is that most of the learning should be within a simulated tactile environment~\cite{ding_sim--real_2020}. This expectation has just been confirmed on several of these tasks using the BRL TacTip~\cite{church2021optical}.


\section{Conclusion}

There are fundamental problems to be addressed in intelligent robotic interaction with complex environments that once solved will open up many application areas across engineering and robotics. One key problem is that there is a huge gap between what is achievable in research laboratories investigating robot manipulation and what is known about human dexterity and our sense of touch. Research on the TacTip aims to bridge that gap as an example of a SoftBOT sensor, combining Soft, Biomimetic, Optical and Tactile sensing.


SoftBOT sensors offer the opportunity to artificially recreate key aspects of the human sense of touch and our manual intelligence. This aim has two interconnected goals: (1)~to advance knowledge of how our sense of touch leads to haptic intelligence from embodying those capabilities in robots; and (2) to improve the intelligent dexterity of robots with accessible robot hardware and software. Reaching human-like levels of dexterity has been the vision for industrial robotics since its roots in the 1950s, and is captured in the earlier origins of `robot' from the Czech word {\it robota} for forced labour. Likewise, using biomimetic touch to achieve that goal has driven developments in robotic tactile sensing since the 1970s. A combination of advances in soft robotics, biomimetic tactile sensing and AI could enable that vision to become reality.


\section*{Acknowledgments}

I am fortunate to have the opportunity to work with many talented PhD students and postdocs at Bristol Robotics Laboratory, including Noor Alakhawand, Kirsty Aquilina, Sophie-Anne Baker, Thomas Cairnes, Alex Church, Pernilla Craig, Luke Cramphorn, Elena Giannaccini, Kipp McAdam Freud, Michele Garibo, Thom Griffith, Anupam Gupta, Fliss Inkpen, Jasper James, Haoran Li, Yijiong Lin, John Lloyd, Fraser MacDonald, Dabal Pedamonti, Nick Pestell, Efi Psomopoulou, Lizzie Stone, Emma Roscow and Ben Ward-Cherrier.

\bibliographystyle{unsrt}
\bibliography{SoftBOTS2021-Review} 

\end{document}